# Efficiency Comparison of AI classification algorithms for Image Detection and Recognition in Real-time

**Musarrat Saberin Nipun**

*School of Computer Science and Technology*
University of Bedfordshire
Vicarage St, Luton, LU1 3JU

**Rejwan Bin Sulaiman**

*School of* Computer Science *and Technology*
University of Bedfordshire
Vicarage St, Luton, LU1 3JU

**Amer Kareem**

*School of Computer Science and Technology*
University of Bedfordshire
Vicarage St, Luton, LU1 3JU

# Abstract

Face detection and identification is the most difficult and often used task in Artificial Intelligence systems. The goal of this study is to present and compare the results of several face detection and recognition algorithms used in the system. This system begins with a training image of a human, then continues on to the test image, identifying the face, comparing it to the trained face, and finally classifying it using OpenCV classifiers. This research will discuss the most effective and successful tactics used in the system, which are implemented using Python, OpenCV, and Matplotlib. It may also be used in locations with CCTV, such as public spaces, shopping malls, and ATM booths.

# Keywords

*Artificial intelligence, classifier, Face detection, Face recognition, python, openCV*

Table of Contents

# Introduction

In this 21st-century, face detection and recognition have become an important and exciting area of research for the students and experts. This is because of its worth in real life and the rapidly growing demands of information and technology for a comfortable and secure lifestyle. Every single day scientists are discovering new technologies to ensure the highest comfort of human life and security in every purpose. Face detection and recognition have become more popular recently because of the invention of active algorithms, a big database of facial images and the invention of performance measurements algorithm of face recognition methods (Phillips et al., 2000). Security systems are more dependent in face recognition software's because compared to normal security systems, biometrical expertise are always providing higher level of security (Ballad, Ballad and Banks, 2011).

Though there are a lot of different options for biometric security like fingerprints and retina scan despite all these face recognition is accepted by all individuals due to its deferential and non-invasive way of identification and validation (*D. R. Patel. ,2008*). For example, in fingerprint recognition a person can get hurt in his finger and cannot be able to access where it required and in worst case, terrorist can cut the finger and use it to access where fingerprint recognition is needed. For this reason, face is the most trusted security system to the individuals.

Many experts used different techniques and algorithms for face recognition system to ensure the improvement and efficiency of the system. Implementation of some of the languages are not that much satisfactory, like java, as the processing speed was very slow and had some complex issues. In this project, python is being used as programming language, and for the standard API openCV library is utilized. This paper is going to detect and recognize movie actor images as its main goal is to apply different algorithm for face detection and recognition and then show the results achieved from them and after that, analysing the different results. Finally, it will provide a conclusion about which algorithm serves better. In other word, this project work can be considered as a research based project as this actually provides information of different techniques implementation and comparisons.

Researchers are trying to solve some methodological barriers that pose a limitation in AI real-time application. Various research has been done in different domains such as Information efficiency (Rejwan Bin sulaiman et al,. 2021), biometric identification (Vitaly Schetinin et al., 2018), Diabetes Prediction (Hassan et al,. 2021), Customer assistant chatbot (Rejwan Bin Sulaiman et al,. 2019), Pneumonia Detection (Amer Kareem et al,. 2022), Credit card fraud detection (Rejwan B.S. 2022). Despite these limitations, researchers are working to gain the ML power to detect images in real time videos.

## Project Background

When choosing the topic for a project I analyze different ideas and finally come up with the most interesting and emerging idea that would certainly be a great resource for the researcher who worked in artificial intelligence. At first, we studied some more research papers and watch tutorials to know more about the face detection and recognition project. While researching, WE could not find any specific paper where all the techniques of face detection and recognition are explained elaborately with examples and provides the implementation information. Moreover, a comparison of all the techniques and their results are not available in one single paper. Though, it takes more time to research about the topic; finally, WE planned for a project which would be more resourceful as this will represent every basic concept of face detection and recognition project and will also provide comparative analysis within the methods.  aim was to build an application-based project which also provide some useful research data, so that, in future who are willing to work with this kind of project, can have a clear concept by reading this paper.

To complete our goal at first, We prepared our project proposal based on face detection and recognition project, where We take the movie actor images to make the research and implementation more interesting. Though WE are implementing the system, but this project can be said as a research project as well, as We are showing all the techniques and their efficiency analysis for face recognition.

Another reason for choosing this topic is the popularity of this subject. For example, previously said that this face recognition systems are used for security purposes, but beside this, social networks, like Facebook also, providing this face recognition service when we upload or tag photos. So, it goes without saying that, now a day's face recognition system is popularly being used in many sectors, and research about this will add an extra value in terms of knowledge about artificial intelligence.

## Project Aim, Scope and Objectives

This proposed system will be able to detect and recognize face of movie actors from images.
For this implementation, one renowned and publicly available movie has been selected. Here, WE selected the superhero movie "X-men" for experimenting our project work. WE used the images of two main actor of that movie for face recognition. One is "Hugh Jackman"- who played the role of "Wolverine", and other one is "James McAvoy" who was the "professor" in that movie. From that movie some of the images which are publicly available in the cloud without any copyright issue, WE choose only those images to be used for our project. Image of these two actors is being used to train the system for face recognition, and also as test image to ensure if the algorithm is recognizing the faces properly. There are few more features include in this project which also includes face detection from live video streaming. Basically, the features of this project include:

1. Detect an actor face from an image
2. Detect all the faces from a group image of actors
3. Detect face/faces from live video streaming
4. Recognize the actor's face from an image

The challenge of this project includes finding out the lacking within the algorithm when the system fails to show the expected result. It can happen if video/image is not properly clear or change in lighting. Another challenge is to find out how the system handled this failure. To compare the highest accuracy rate within the algorithm is also a big challenge.

Though face recognition is a very demanding research topic in current days, but still there is a lot of scope to research in this subject and finding more efficient implementation method. This project will be helpful to those who has interest in face recognition but beginner in this field. As this paper is using OpenCV which is the most popular and efficient API for face detection and recognition, and also implementing the algorithm using one of the demanding programming language Python, it will help those beginner students who are planning to make there career in deep learning/machine learning. Moreover, this paper not only showing the concepts and techniques of face recognition, but also it simultaneously provides the idea of implementation and a clear comparison analysis to decide, which algorithm is needed to be used in which situation. Beside this, this system will show the implementation using movie actors face, which will motivate the young learners about face recognition. The complete system can be used as a component of a security system where face detection and recognition are needed such as, any shopping mall, retails shops,

ATM booths, Cars, banks and even in homes where there is CCTV activated. So, day by day use of face recognition system will increase and an easy and efficient system will always be on demand.

# Literature Review

When starting a project, one of the crucial requirements is to gather knowledge about other works with the same or similar subject. It helps to find out the difficulties and drawbacks of previous work which need to be improved in future research. It also provides the information about what technologies are currently available to implement the system and the brief idea of which one is more efficient and can be improved. Literature review is the broader meaning of deep analysation, finding of logical proves, planning of robust solution and reaching in efficient conclusion.

The face recognition process has a very rich background history. In need of an efficient face recognition system lots of research has been done in past few years. The goal of those research was to introduce a more accurate face recognition system with less time consumed and test and analyse the proposed system. Also, to compare various face recognition algorithm which were previously tested by expertise.

The platform of face recognition algorithm pioneered by Service and Kirby (lata,Y.V. et al, 2009) who followed the Eigenfaces approach for face recognition system. (Ali, E.S et al. , 2021) Sirovich and Kirby implemented a software which can identify the head of a subject and then compare the characteristics with other known faces it already has in its database and then recognizes it.

Another face recognition algorithm was invented using independent component analysis, which is done by Bartlet, Movellan and Sejnowski (Bazama, A. et al, 2021) Two types of algorithm method has been in consideration for face recognition. One is PCA which means Principal Component Analysis and another one is ICA: Independent Component Analysis. PCA is based on the idea of pixels where important data of an image is stored in pairwise relationships (Bhattacharya, A., 2021). On the other hand, ICA is focused on the logic that some important data can contain high order statistics. Finally, researcher Maryam Mollaee and Mohammad Hossein Moattar proposed an improved ICA model which shows better accuracy in face recognition system (Sujay, S.N.et al, 2019).

The previous Eigenface method had a problem that it cannot identify face in different illumination condition. To solve this problem, in 1996 using Linear Discriminant Analysis a new method was developed for dimensional reduction which is called Fisherface (Kirana, K.C. et al, 2018). A face detection algorithm was proposed by Viola and Jones using HAAR Cascade and AdaBoost (Smashing Magazine, 2019).

**Face Recognition from Movie Actor Image**

Most recently Another research has been done by Remigiusz Baran and Filip Rudzinski on 2016 (Ballad, Ballad and Banks, 2011), which is quite similar to this one, that was also recognition of faces from the Movie Actor's image or video. Though the plot of the research is same, but this research took a different approach while implementing the system. That was based on content discovery and delivery platform known as the IMCOP system (Baran and Zeja, 2015). IMCOP is a SOA driven platform where the intelligent utilities which are needed for the face recognition system is implanted as REST (Representational State Transfer) web services (Mitchell et al., 2008), which is actually called IMCOP's Metadata Enhancement Service (MES). Every single MES service is processing a different task.

# Research Methodology

There are a few methods that are being used for face detection and face recognition. In this project, as we are using OpenCV, we will implement all the renowned techniques for face detection and recognition and from the achieved result we will compare the accuracy of the techniques. The methodologies for face detection and recognition are discussed in separate sections in this report.

## Research Methodology for Face Detection

Face detection has achieved more popularity due to its invention of real-time applications. Still, there is a lot research going on in face detection, to make the algorithm more efficient and accurate. Face detection is not at all an easy task because it always includes faces and it is hard to detect all the variances of a face appeared. For example, pose variation- like smile, no-smile, with hair, no hair, with beard, no beard etc. To handle these things choice of tools is always important and so, we are using openCV here. OpenCV is a computer vision and machine learning software library which is mainly aimed at real-time computer vision and is open for all to use (Opencv.org, 2019). This library has more than 2500 algorithms which include machine learning tools for classification and clustering, image processing and vision algorithm, basic algorithms and drawing functions, GUI and I/O functions for images and videos (Opencv.org, 2019). OpenCV library has some built-in pre-trained classifiers for detecting faces, eyes and smiles etc. Specifically, for face detection, openCV has two pre-trained classifiers:

1. HAAR C **(Conti, et al., 2013)**
2. LBP Cascade classifier

We are going to use them in our project and the detail of these classifiers are discussed in this report.

### HAAR Cascade Classifier:

This algorithm is invented by Paul Viola and Michael Jones. (Viola and Jones, 2001). In this algorithm, a function is trained with different images. This function is called as cascade function. Images can be positive and negative. Images which include face are called positive image and which are without face is called negative image. This approach is mainly based on machine learning.

This algorithm basically has four stages as follows.

1. **Selecting Haar feature**: Assume that we have an input image. To get the facial features large number or HAAR features are needed. That starts by extracting Haar features from each image. The windows which are being used to subtract the pixels of a face are shown in below image:

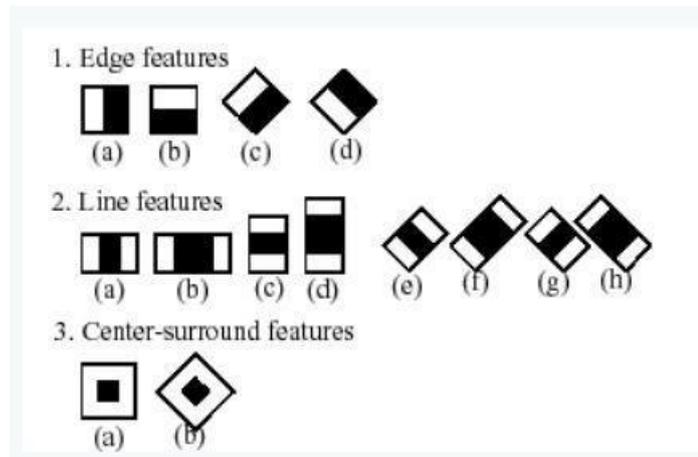
*Figure 2: A couple HAAR like features (SuperDataScience Team, 2017)*

To estimate a single feature one by one each window has set on top of the image. This feature takes the sum of the pixels under the black part of the window and do the same for the white part of the window. Then subtract the summation value gained from the white part of the window, from the summation value gained from the black part of the window. By doing this subtraction this feature gets a single value. After that, to calculate enough number of features each window are set on all feasible locations of the image depending on the size of the window. Below picture shown the example of this features.

2. **Creating an Integral Image:** To make the algorithm fast this method creates an integral image. Before this, excessive computation occurred while performing operations on all pixels. This integral image minimizes the computation rate only in four pixels.

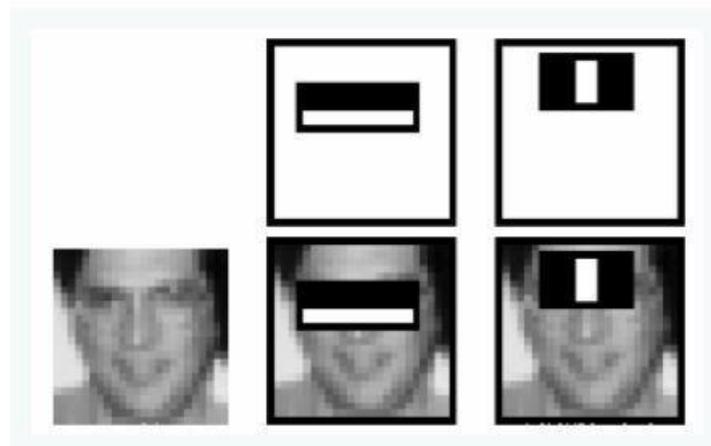
*Figure 3: Different stages in visualization (SuperDataScience Team, 2017)*

The above image is showing the extraction of two features. The region surrounding eyes are darker than the region of the nose and cheeks, the first stage of features concentrate on this attribute. At the second stage, the features are based on the attribute that the bridge of the nose is lighter than the eyes.

3. **Adaboost:** The features which were calculated almost all of them are inappropriate. Such as, in case of cheek, used windows are inappropriate as there are no dark or light part to compare with other regions of the cheek. All the areas look like the same. So, all the features cannot be used for the classification purpose. Here, adaboost is being used to categorize the relevant features. Adaboost is a face detection training procedure. It chooses only the features which are used to improve the accuracy of the classifier.
4. **Cascading Classifiers**: In this stage only, the relevant features are used to detect a face. This algorithm uses the cascade classifier concept which can recognize that every area of the image is not related to face and so it doesn't apply all the features on the full area of the image, rather it group the features to use in different stages of the classifier where needed. To find the face it applies each stage sequentially. As a result, if the classifier fails in any stage it rejected that region for any future iteration. The classifier only allows face areas to pass all the stages.

## LBP Cascade Classifier

LBP means local binary pattern cascade classifier which is actually a visual/ texture descriptor classifier. This classifier is also needed to train with images. In this classifier every feature is extracted and together they form a feature vector. This feature vector is used to categorize a face from a non-face. The steps of this classifier are described below:

1. **LBP Labelling:** In this classifier at first every pixel of a picture is allocated with a string of binary number which is called label. At first it divides the training image into some blocks.

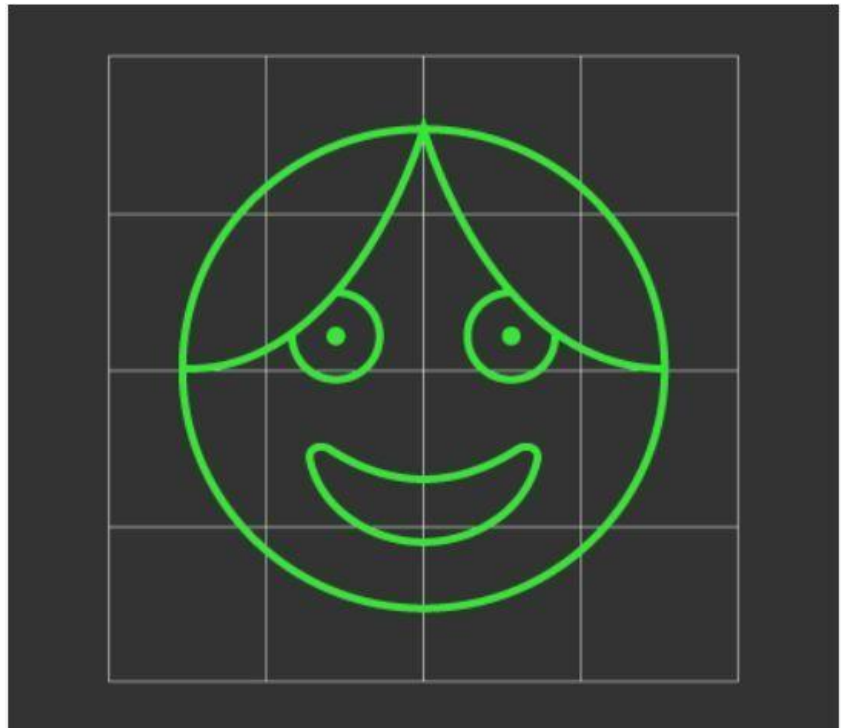

*Figure 4: LBP windows (SuperDataScience Team, 2017)*

When considering the blocks, LBP focuses in the pixel which is at the centre of the window and also at the same time, it focused at 9 pixels which is a 3X3 window. After that, within the 3X3 window, it starts comparing the pixel value defined for the centre with each of the neighbouring pixel value. Then, it defines the binary value 1 for the neighbouring pixel value which are greater than or equal to the value of the centre pixel, otherwise it defines them to 0.

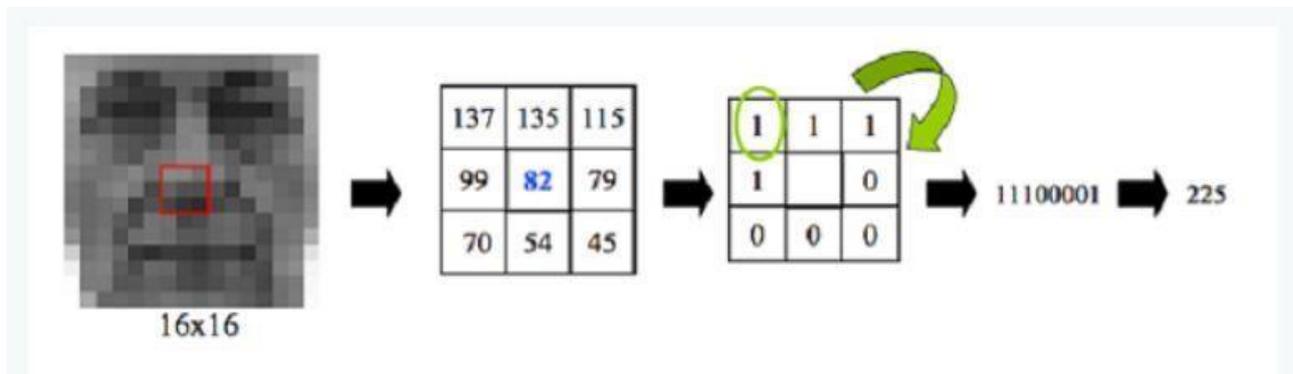
*Figure 5: LBP conversion to binary (SuperDataScience Team, 2017)*

Next, it forms a binary number by reading the updated pixel values in a clockwise order and converts that into decimal. It uses that decimal number for the centre pixel as a new value. It happens with all the pixels in a block.

2. **Feature Vector:** To create the feature vector the given picture is divided into different blocks. A histogram of label is built for all these blocks. Finally, a large histogram is made by connecting all the blocks histogram and made the feature vector.

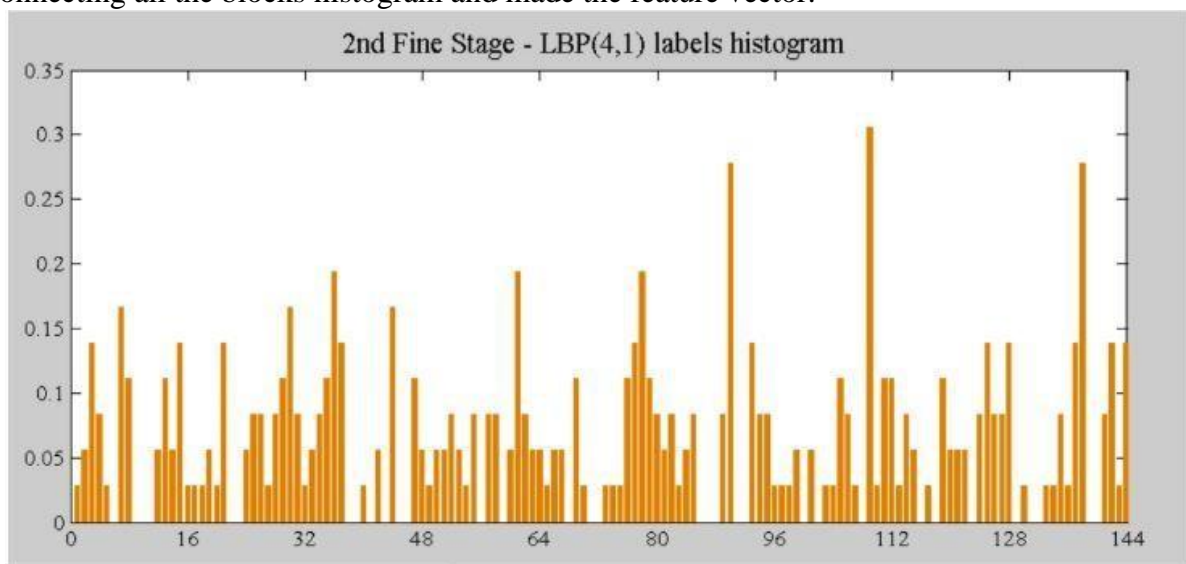
*Figure 6: LBP histogram (SuperDataScience Team, 2017)*

3. **AdaBoost Learning:** Similarly, like Haar cascade, to erase inappropriate data from feature vector adaboost is used which helps to create strong classifier.

4. **Cascade of Classifier:** The features achieved by adaboost creates the final cascade classifier. The image blocks are evaluated from easier classifier to strong classifier. Classifiers which fails to pass a region in any stage are rejected for future iteration. It is only possible for the facial region to pass all the stages of the classifier.

In this report, we will use both of the classifier, compare the robustness and will analyse the accuracy rate for both of them.

# Research Methodology for Face Recognition

Face recognition system often compared to the process of how our brain identify a face. For example, when we meet with one of our friends, by seeing his face we instantly know his name and detail. This is because, when we met with that person for the first time, our brain took the necessary information about that person. For example, at first our brain looks at his face, eyes and nose and when he introduced himself by saying his name, it stored into our mind. Actually, our brain is learning or training about that person on that time. Next time we see the same person we can easily recognize him. The interesting fact is, when we introduce with new person, at first, we notice the face of the person and in future we easily recognize the person by seeing him or a picture of him because of memorizing his face. That is how our brain is doing the face recognition. Almost the same case happened with the face recognition system.

When it comes to face recognition system, it follows almost the same process as our brain do. In our project, as we are using openCV, it is easier to code that system as well. The steps of face recognition systems are discussed below:

1. **Face Detection:** At first, the system looks at the image and find the face into it.
2. **Training Data Gathering:** Then it extracts the important and unique characteristics like, eyes, mouth, nose etc to memorize the face so that it can differentiate this face with another face.
3. **Training of Recognizer**: Then it trained the face recognizer the names of each faces by feeding the face data, so that it can compare the unique features with all the features within all the people whom the recognizer knows.
4. **Recognition:** Then it feed the new faces of the persons of which the recognizer is already trained and see the results.

## OpenCV Face Recognizers

In this project, as we are going to use openCV face recognizers, we are going to describe three built in face recognizers in this report which we will use in our project. The name of these three recognizers are:

1. EigenFaces Face Recognizer
2. FisherFaces Face Recognizer
3. Local Binary Patterns Histograms (LBPH) Face Recognizer

## EigenFaces Face Recognizer

This eigenface method is invented depending on a statistical approach by (Turk & Pentland, 1991) (Jaiswal, Gwalior, & Gwalior, 2011) (Morizet, Ea, Rossant, Amiel, & Amara, 2006). The main technique of this method is to extract that component which causes the most variation in an image. This component is called principal component. Therefore, another name of this technique is principal component analysis. This is also called a holistic approach, and the prediction is based on the entire training set. If the images are from two different classes, this method could not provide any specific solution for these images. Here, a class means a person. This is actually a machine learning technique where grayscale pictures are used for training the recognizer.

When we look at any face our brain automatically determine that all the parts of a face are not important to remember, and it extracts the most unique and relevant data which it will use to recognize the face in future. This algorithm also considers the same process. This recognizers at first looked at all the persons every training image and focused on that part of the face with maximum change. Then, it extracts the important and useful features from the face and read the distinct features like eyes, nose, mouth etc. It gathers the information of the variance of the face and reject that part of the faces which is not important to memorize. As an example, it focused from eyes to nose or nose to mouth as there is a significant change. When it reads multiple faces, it memorizes these distinct features to recognize the face in future and differentiate the face with others. In this way it not only memorizes the important data face but also as it is rejecting the unimportant contents, it helps to save memory. The important components are called as principal components which is shown in below image.

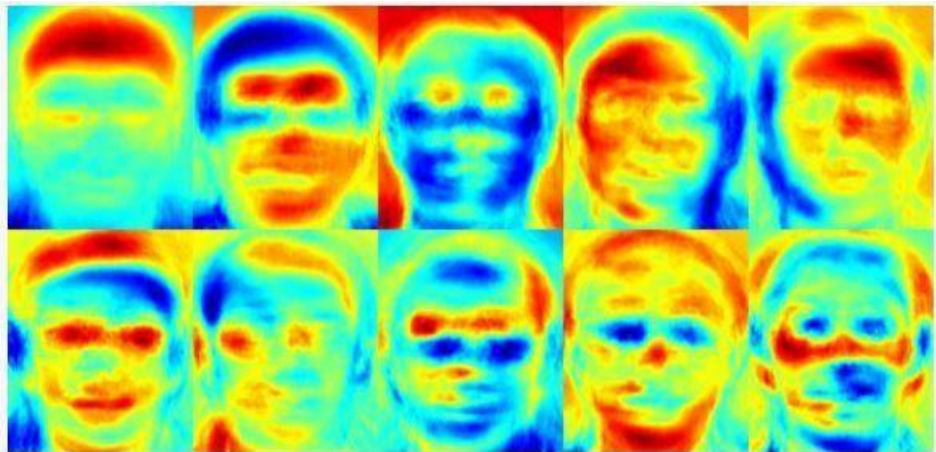

*Figure 7: Eigenfaces face recognizer principal components (SuperDataScience Team, 2017)*

From the picture, it is shown that the principal components which is shown in the picture presents eigenfaces. This algorithm trains itself by extracting these components shown in the picture. This algorithm keeps separate record of principal components for each of the faces which means it knows which principal components belongs to which face.

In this algorithm it is essential to have same lights in every image and also the eyes need to be match in every image. Another important thing to keep in mind that the number of pixels of every image should be same and should be in grayscale. As an example, we can assume that a n x n pixels image where each row is joined in order to make a vector. As a result, it makes a $1 \times n^2$ matrix. Here, all the images which are going to be used for training data set is stored in a single matrix which leading to a column matrix equivalent to the number of images. Then, to get an average human face the matrix is being standardized. Unique features of each face have been calculated by subtracting the average face from the image vector created for each image. The difference of each face from the average face is represents by each of the column in the resulting matrix. The complete process is shown in below image:

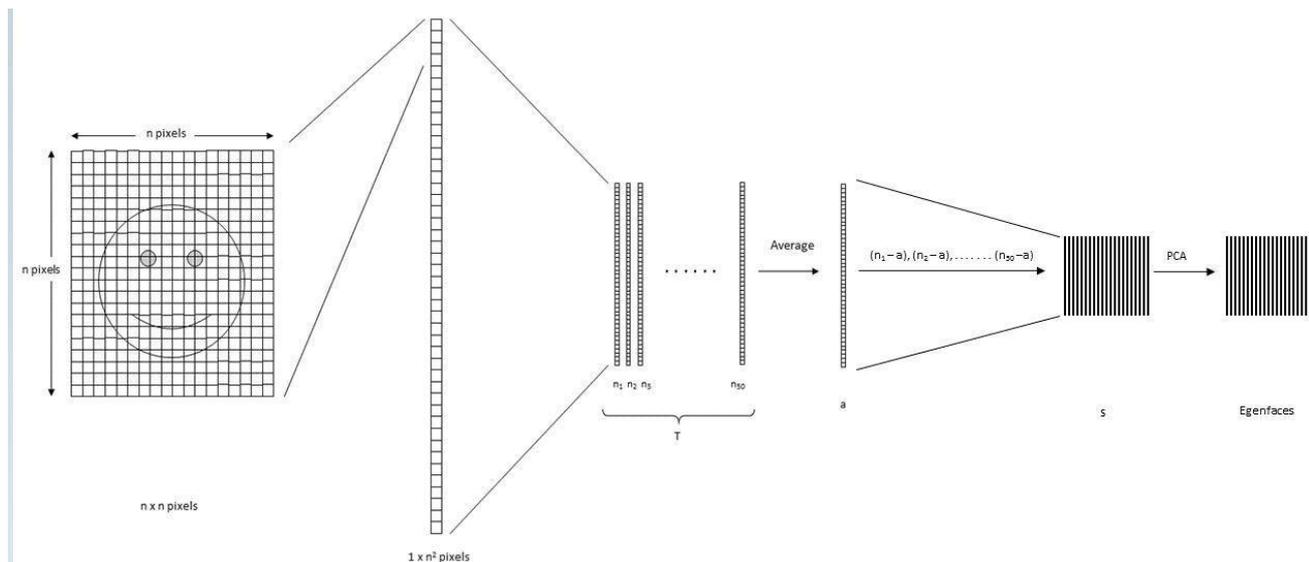
*Figure 8: Reordered pixels of the image to perform calculation in Eigenfaces (Dinalankara, Lahiru, 2017)*

After this step, it is needed to compute the covariance matrix from the result matrix. From the training data, to find out the eigen vectors using principal component analysis, eigen analysis has been performed. In the result matrix where the covariance matrix has been found diagonal, there it has the highest variance which is considered as 1$^{st}$ Eigen vector. Then, the next highest variance is the 2$^{nd}$ eigen vector and this is 90 degrees to the first one. Then, the next highest variation will be the third one. This process is going to same as this for further vector. Here, each column is assumed an image and matches face, and these faces are called eigenface. To recognize a face, it takes an image as input then resized it to match with the same dimension of that eigenfaces it already retrieved. When this algorithm got a new image to feed, it repeats the same process on the image and compare the components with the list of the components it already stored while training and search the best match with the given components and return the recognition result.

This principal component analysis causes problem with large numbers. Another important fact to this algorithm is, it also considers the light and shadow as an important component when recognizing a face. Even the same face cannot be recognized if it analysed under diverse lighting condition as this algorithm undermix the values while calculating the distribution and failed to be classified efficiently, which causes a problem while matching the features.

## FisherFaces Face Recognizer

The next algorithm is fisherface recognizer which is a developed version of the previous one.
We discussed the process of eigenface face recognizer earlier, but it has also some lacking. For example, in eigenfaces if the image has some slight changes, such as, changing in lighting or shadow, it overlooked the other images, but this slight change is also an unimportant feature to consider. At the end, maybe it creates some features from the outside source which is not important to consider and for this it may unable to determine actual face features.

Fisherfaces (Jaiswal & al., 2011) (Morizet & al., 2006) (Belhumeur, Hespanha, & Kriegman, 1997) algorithm is though similar to eigenfaces algorithm but it has some slightly different technique than eigenfaces method which made it more developed. This algorithm also uses a holistic approach. It also uses the principal component analysis technique, but it changes that technique a bit as it takes consideration classes. As it is described earlier that eigenfaces cannot differentiate two pictures from different classes during the training session, but Fisherfaces uses a method to differentiate two pictures from different classes. For this, it uses Linear Discriminant Analysis method. The goal of using this method is to minimize the variation within a class by comparing the variation between the classes. To do this, it not only uses total average of faces but also takes the average of per class.

Eigenfaces algorithm extract important features which presents all the faces of all the persons, here, Fisherface algorithm only extract the features which will help only to differentiate one person from all the other person. This way it brings it out the differentiation features of a person from other and the features are not dominant on other features. The process is shown in below image.

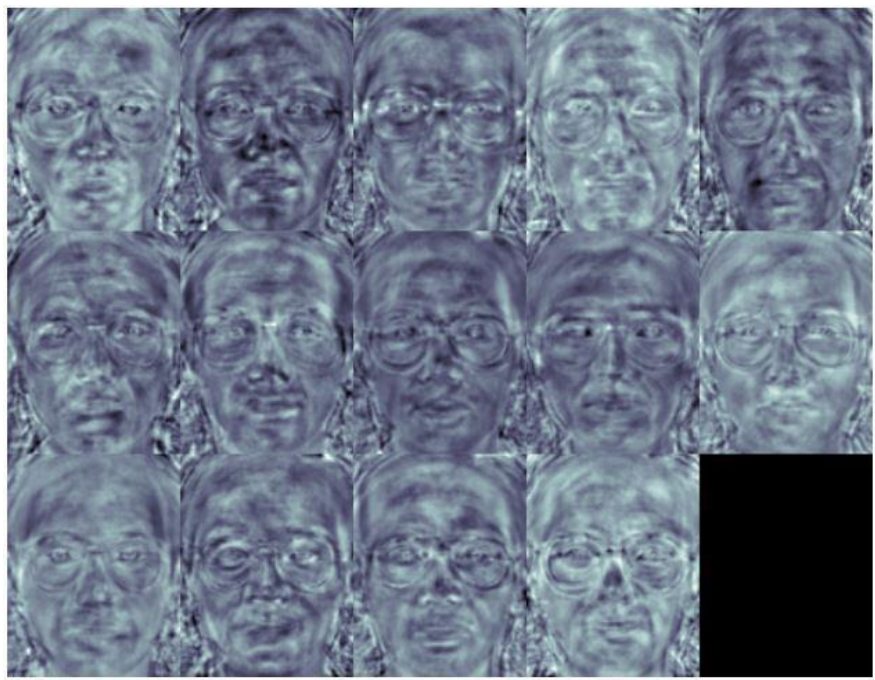

*Figure 9: Fisherface algorithm (SuperDataScience Team, 2017)*

The above picture shows the extracted face features which is called fisherfaces. Like the eigenface algorithm fisherface has some lacking too. If there are any slight changes in images from any outside source, it could create confusion within the features which will hamper the accuracy rate. The outside source can be the change of light and shadow as described earlier.

**Local Binary Pattern Histograms (LBPH) Face Recognizer**

The last algorithm we are going to discuss in this section is the Local Binary Pattern Histograms (Ahonen, Hadid, & Pietik, 2004) (Mäenpää, Pietikäinen, & Ojala, 2000) (Wagner, 2011) face recoqnizer. The goal of this algorithm is to work with 3X3 pixel blocks. The difference of this one from the others, that we have already discussed is that, this one is not a holisitic approach though it requires grayscale images for training as well.

Previously, we discussed that both the features are unable to present proper accuracy rate in face recognition due to effect of light and shadow and in real life scenario it is impossible to confirm perfect light conditions in every image. To overcome this lacking LBPH face recognizer algorithm improved its functions.

LBPH does not look at the whole image rather it searches for the local features of an image. Like the LBPH face detection techniques here this algorithm compared each pixel with its neighbouring pixel and search the local construction of an image. It shifts an image by taking a 3x3 window. It compares the pixel of the centre of that image to its neighbouring image while shifting each local part of an image. The neighbouring pixels which has value lower than the centre value are marked 1, otherwise 0. Then, the algorithms read that binary values maintaining the clockwise order under 3x3 window and make a binary pattern like 11100011 which is the local features in some areas of the image. Then, the same process iterates in the whole image and the algorithm gets a list of local binary patterns.

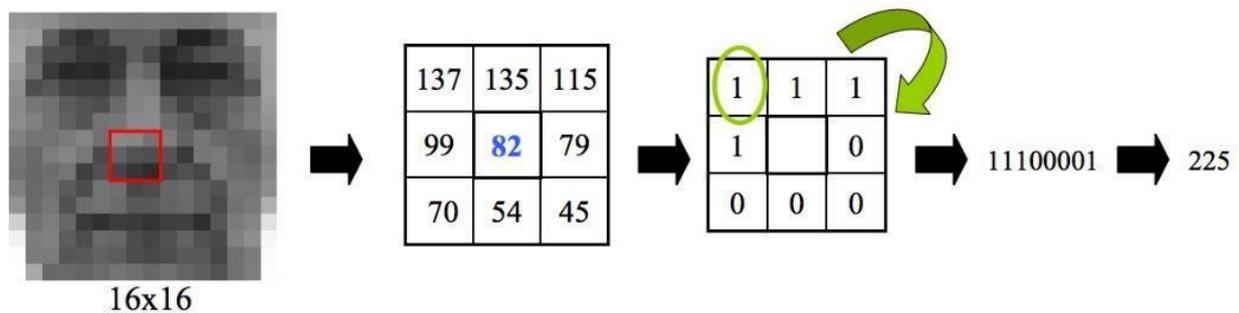

*Figure 10: LBP Labelling (SuperDataScience Team, 2017)*

After that, the histogram part comes which completes the algorithm for the face recognizer. After getting the list of local binary patterns, it is needed to convert them into decimal number, which is shown in above picture. Then, we can be able to make a histogram with these decimal values. Below is the picture for sample histogram.

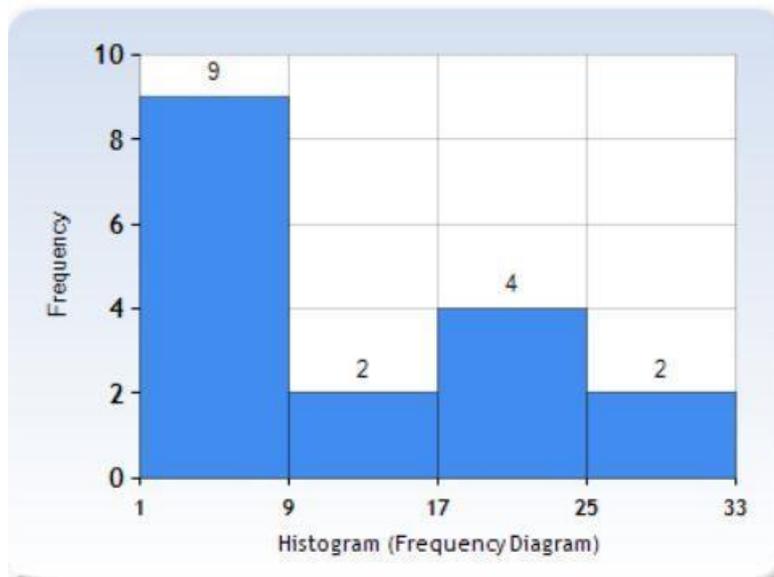

*Figure 11: Sample Histogram (SuperDataScience Team, 2017)*

Finally, in the training data we will find one histogram for every face image which means if we have 100 images for training data at the end LBPH face recognizer will find 100 histogram for each of the faces and save them for face recognition in future. An important feature of this algorithm is it also can be able to track which histogram belongs to whom.

Finally when we do the test data face recognition, this algorithm makes a new histogram for that test image and then it compares that histogram with all the other histogram it already has from its training data set and find the best matched histogram, then it return the person label associated with that histogram.

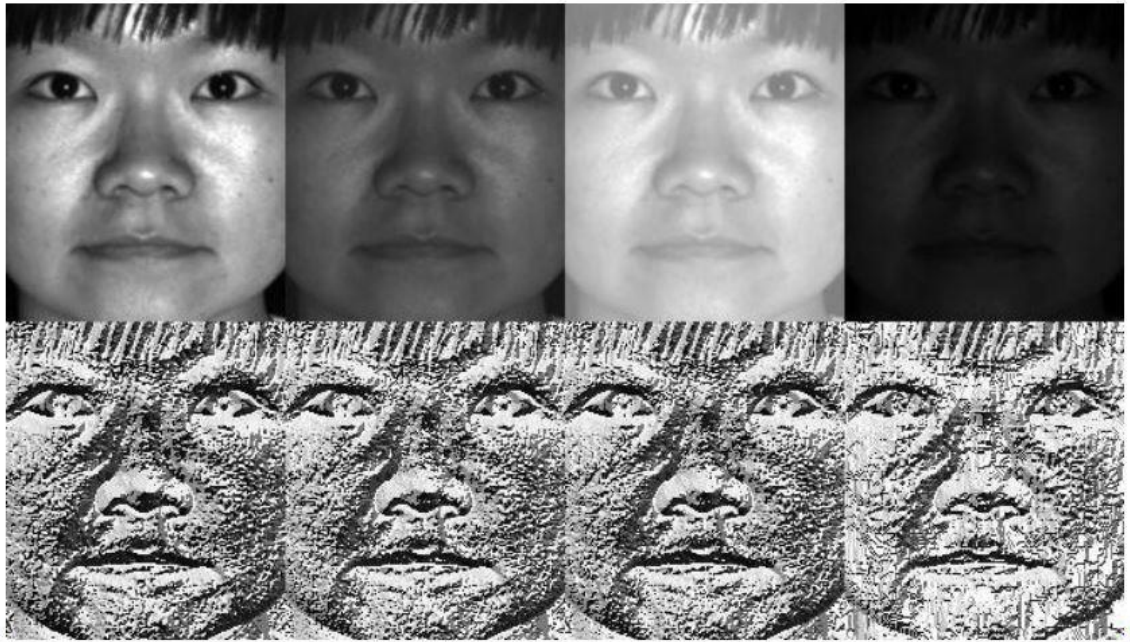

*Figure 12: LBP faces (SuperDataScience Team, 2017)*

Above picture shows a list of faces and the LBP faces made by LBPH face recognition algorithm. It shows that with the change of light the histogram is not affected, so it can be assume that this algorithm ensures highest accuracy level in face recognition. Now, we are going to see this by implementing all the algorithms and comparing their results in next sections.

## Face Detection Development Methodology

For face detection as there are two built in classifier in openCV, this paper will discuss the development methodology for both of the classifiers as we are going to implement both of them. Let's consider the HAAR cascade classifier first for face detection.

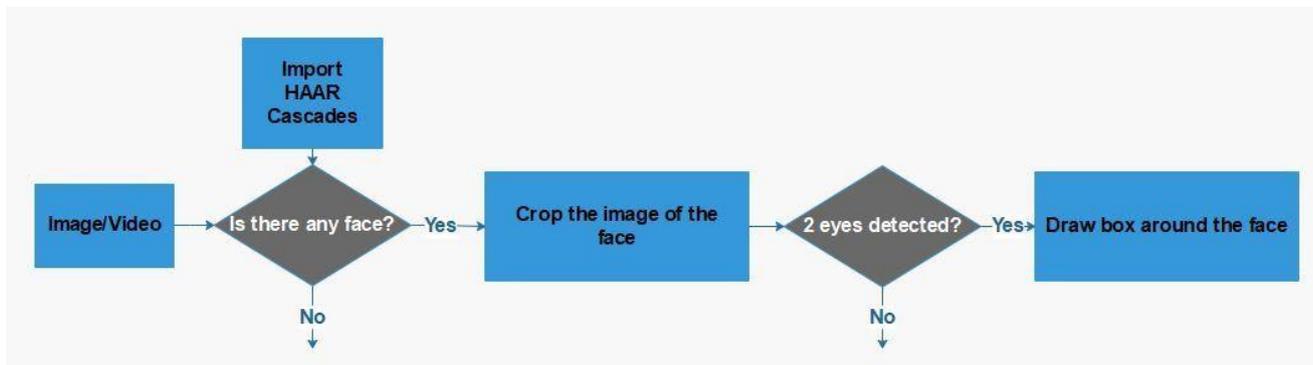
*Figure 14: Flow diagram of face detection development methodology (Dinalankara, Lahiru, 2017)*

This process flow diagram is showing how the system will collect data and respond. First, I will use HAAR cascade classifier and then gather the result. Then, I'll follow the same flow diagram with LBP cascade classifier. At first, the system will take an image or video and also import the HAAR cascade classifier script. If the image or video includes a face, the system will crop the image of the face and find the eyes. When it found two eyes, the system will draw a box around the face which proves that face has been detected.

## Face Recognition Development Methodology

Like the face detection methodology, for face recognition, this paper is going to describe the methodology for three of the classifiers for face recognition. There are three same steps we are going to follow to develop our face recognition project with three of these different algorithms. The steps are-

1. **Collecting the ID for the images.**

   To collect the images, simply we need to take some pictures of a person and then it needs to be cropped and resizez which can be done by any photo editing software. As, Eigenface and fisherface algorithm can not give accurate result if all the pictures are not same size. As this process is time consuming for huge number of images this process can be automated by an algorithm which collects huge number of images with different expressions. The flow chart of this algorithm is shown in below image.

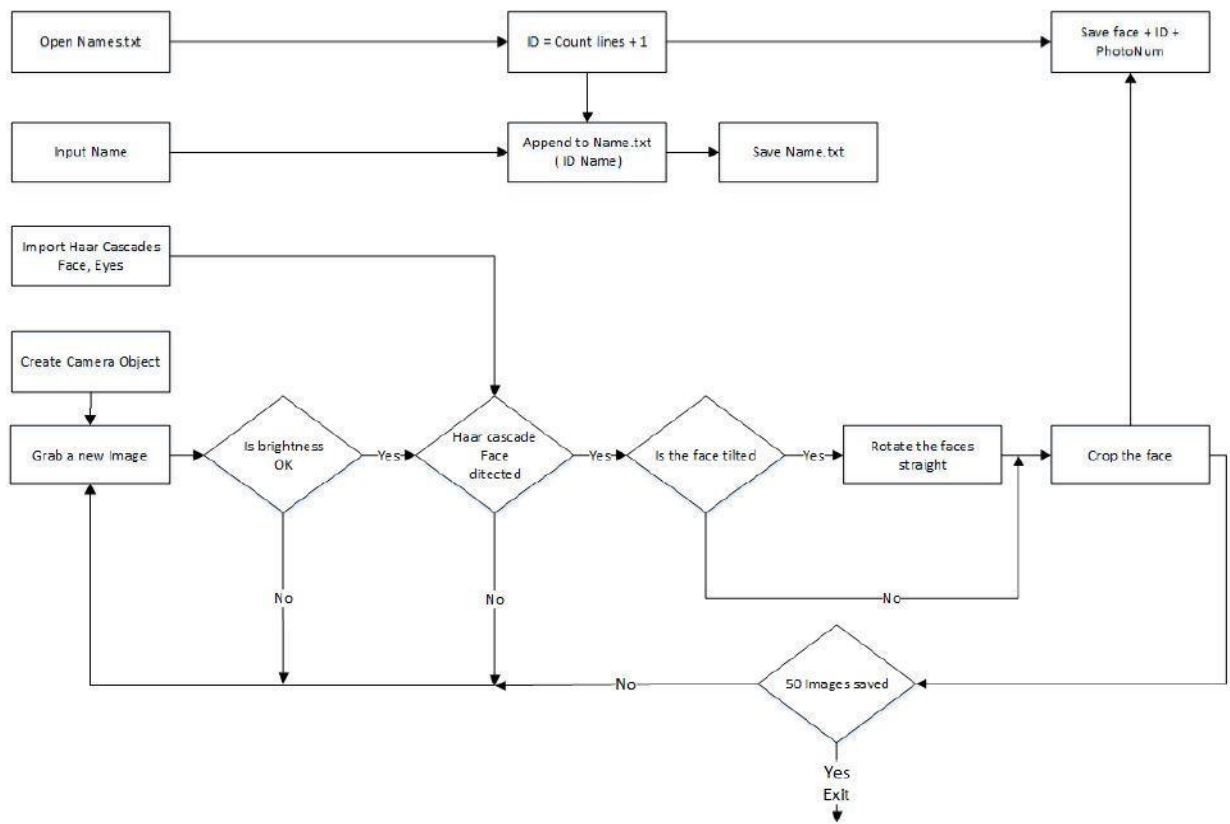

*Figure 15: Flow diagram of image collection process (Dinalankara, Lahiru, 2017)*

At the starting of the application, it takes a request to enter a name to store with an ID no in a text file. At the first half, the face detection system going to be start. Before it takes the pictures, this algorithm looks at brightness level and check if the faces are visible, if it is then it analyses the positions of eyes and rejects the pictures where face is not visible. This algorithm also corrects the position of the head if it is tilted. Then the image has been cropped and saved in a file name using the id so that it can be identified later. This function is run within a loop until it collects all the images. This algorithm we are going to use because of eigenface and fisherface methods, as they are very sensitive about image size, brightness and changes of facial expression. This algorithm also made the collection process faster and efficient.

2. **Train the classifier**

After collecting the images, the next step is to create face recognizer objects using the face recognizer class and train the classifiers. Each recogniers will take some parameters which are described separately.

**cv2.face.createEigenFaceRecognizer()**

1. It can take number of components for the PCA analysis and by opencv documentation it has been said that 80 can give successful result.

2. It can take threshold while recognising a face to give the accurate result. If the distance of the face is above the threshold the function returns a -1 which can be used to determine that the face is not recognized. **cv2.face.createFisherfaceRecognizer()**

1. To create the with LDA the number of components can be determined by the first argument.

2. The threshold is similar to the previous one.

**cv2.face.createLBPHFaceRecognizer()**

1. To build the local binary pattern the radius from the centre pixel need to be calculated.
2. To create the pattern number of sample points is requires, which sometimes makes the computer slow.
3. Number of cells is needed to create in X an Y axis.
4. The threshold is similar to the other two

After creating the recognizer object, then to train the reconizer, FaceRecognizer.train (NumpyImage, ID) function is used. Though, we resized the image, but it is not necessary for LBPH. It is only necessary for eigenfaces and fisherface method. Here, using the same algorithm WE trained three of the recognizers which is very convenient. The algorithm is shown in below flow diagram.

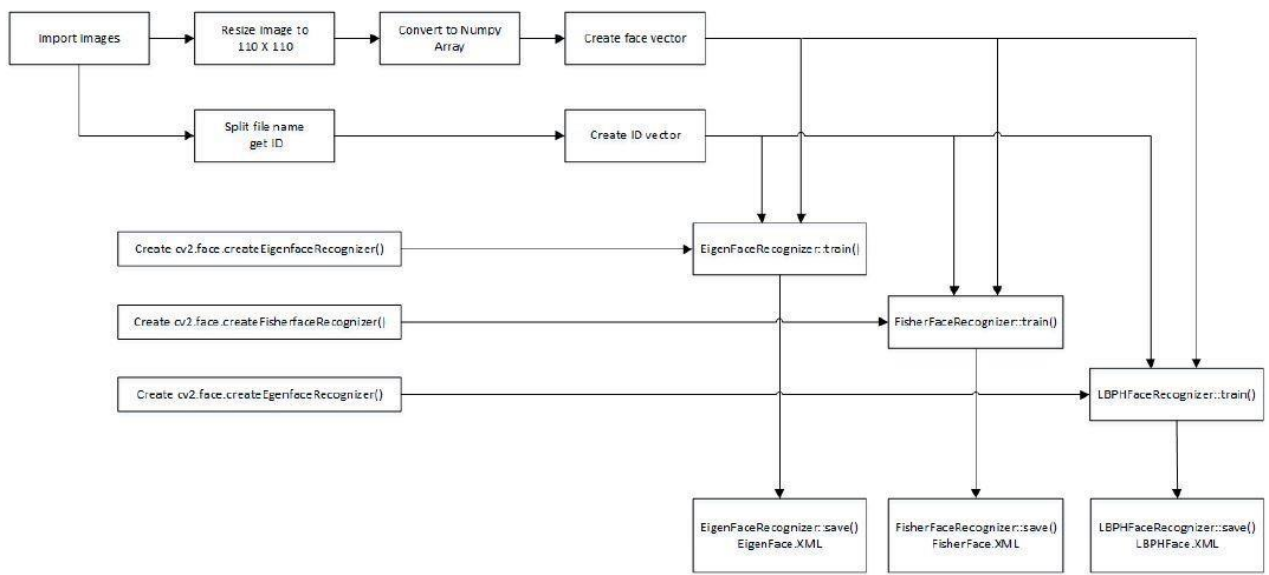

*Figure 11: Flow diagram of training the classifier (Dinalankara, Lahiru, 2017)*

### 3. The face recognition Technique:

The face recognition technique includes matching all the features of the test image with the training image features and predict that face. We will be going to use the same algorithm for the recognition process for three of the methods.

Below is the flow diagram of how the HAAR cascade classifier will be used for face recognition.

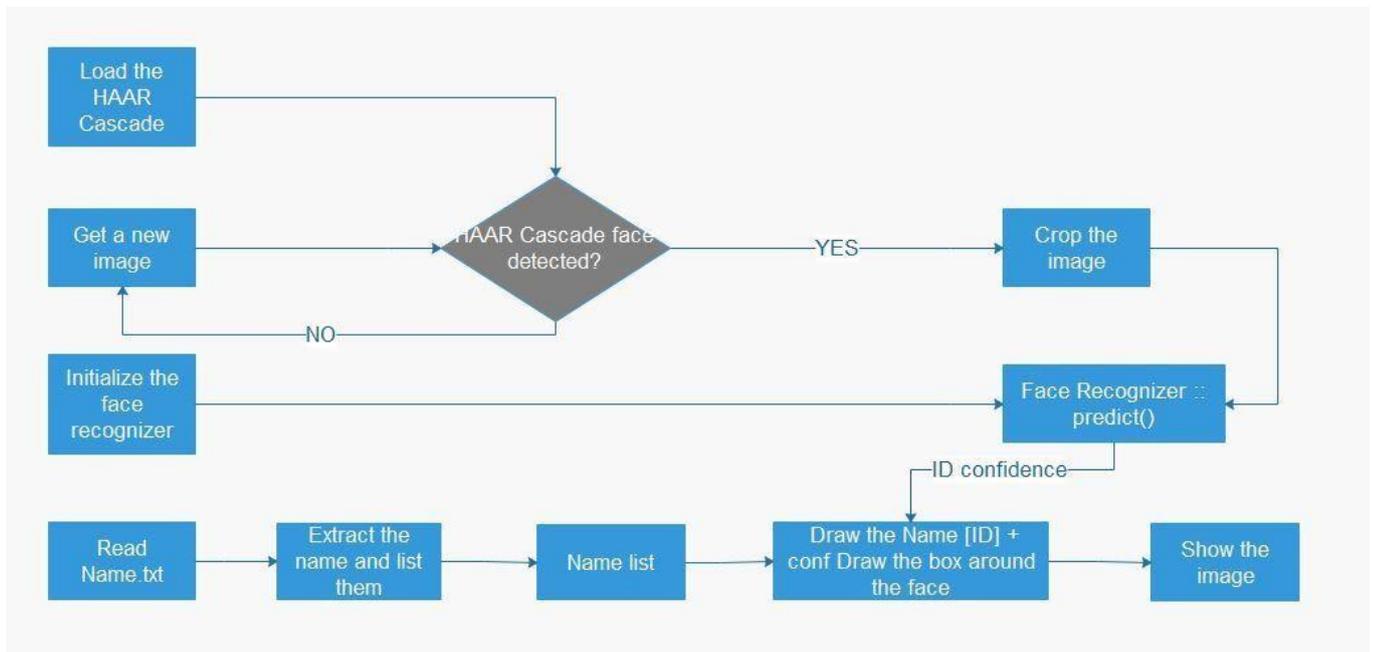

*Figure 17: Flow diagram of face recognition development methodology (Dinalankara, Lahiru, 2017)*

The system will take a new image and load the HAAR cascade classifier at the same time. A face recognizer predict function has already been initialized to predict the face with the help of HAAR cascade classifier. When that classifier detects a face in that image, it will crop the image and run the face recognizer predict function. The name of each faces is already extracted and listed. So, this face recognizer predict function draw the name of the given image and also draw the box around the face and return the face image after recognition. Though, the above flow diagram shows the HAAR cascade classifier face recognition script, but in this project three of the classifier will be used using the above flow diagram.

# System Implementation

While finalizing the project ideas, we researched about various software and languages used for face detection and recognition, so that WE could choose the best one for our project. There are some more software like visual studio and programming languages like c#, java etc, which are popular for coding face detection and recognition but after analysing WE found out that openCV is providing the biggest library for face detection and recognition which makes this process easier than before. To implement the library gifted by openCv WE realized that programming language Python is the best to use, as it implements the techniques in an easier way and also user friendly. No hardware tools were needed in this project implementation but few more dependencies WE add to implement the project which makes process simple to use.

For face detection implementation below software are used-

1. OpenCV 3.2.0
2. Python V3.5
3. Matplotlib 2.0

For Face recognition implementation below software are used-

1. OpenCV 3.2.0
2. Python V3.5
3. Matplotlib 2.0
4. NumPy

Before starting the coding part, first, WE installed the above software. Though, Matplotlib was optional to install but WE install it to see the results in an organized manner. On the other hand, while implementing face recognition system, to make the computing easier WE used the NumPy array to feed data as input to OpenCV function as it contains a powerful implementation of N-dimensional arrays (SuperDataScience Team, 2017).

The coding part start with importing the required modules.

**Required Modules**

1. cv2: This is an openCV module which is being used for face detection and recognition.
2. os: This is a Python module used to read training directories and file names.
3. Matplotlib: This has been used to import time libraries so that we can compare the speed of both classifiers for face detection
4. NumPy: It converts python lists to numpy arrays which will be needed by the face recognizer to recognize the face.

# Face Detection Implementation

In this project, WE have implemented the face detection for 3 kind of features. They are-

1. Face detection from single image
2. Face detection from group image
3. Face detection from live video streaming.

Here, at first, WE am going to explain the face detection using HAAR cascade classifier for image and group image. After that, WE will explain the face detection implementation for live video streaming.

## HAAR Cascade Classifier Using

At first, WE create a folder in the name of the project "face detection and recognition". Inside this, WE create another folder named "OpenCV files" where WE stored the openCV XML classifiers which I'll going to use for face detection and recognition. At the beginning of our python coding, first WE load the HAAR Cascade XML classifier. There is a recognized class in openCV which is cv2.CascadeClassifier. This class takes the HAAR training image and then load it.

Using openCV classifier, when trying to load any image, by default it loads the image into BGR colour as OpenCV do many of the operations in grayscale. Then it shows the gray image. First, WE implement the face detection algorithm for HAAR Cascade classifier.

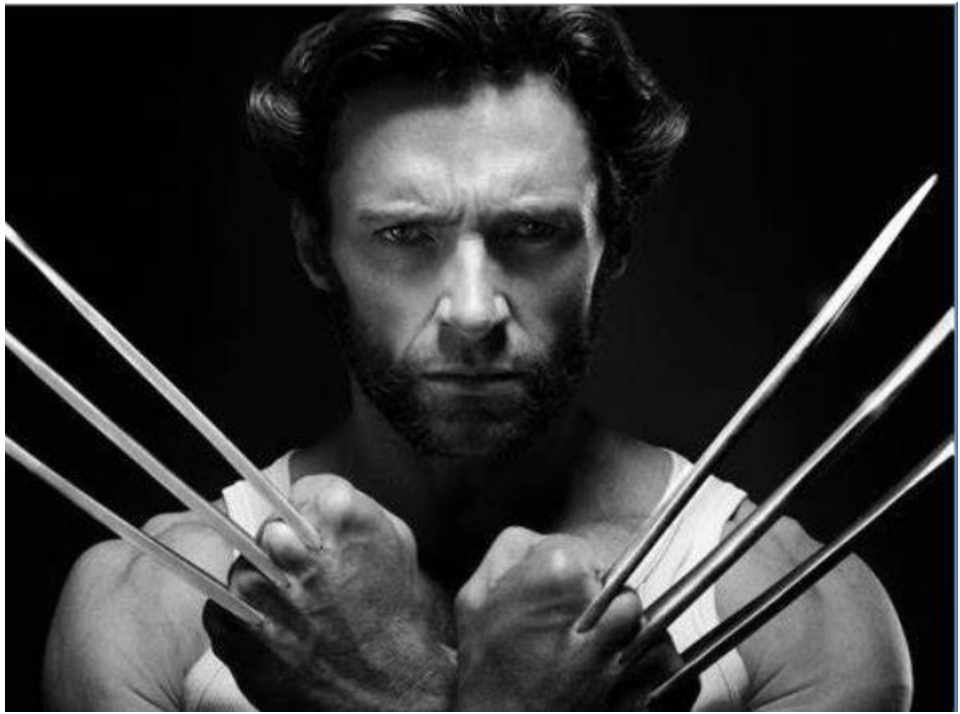

*Figure 18: Converted grey image by openCV*

The loaded cascade classifier is being used to detect the faces. This classifier has a built-in function within it which is called "detectMultiScale" function. Here, to find the faces in the image WE use detectMultiScale functions. This is a general function, which is being used to detect object. Here, WE use this function to detect faces in a test image and it has some parameters which WE pass through the function. The parameters include:

- **Image:** This function is called for detecting the face, so the first parameter is the colour scale for the image, e.g: gray

- **scaleFactor:** Scale factor parameter is being used to determine the distance of faces; as some of the faces can be near to the camera which will appear larger, and the faces which are in back can be shown smaller. The scale factor reimburses for this.
- **minNeighbors:** To detect an object this detectMultiScale function used a moving window and it determines all the objects into an image until it find the face. This parameter defines the other objects.

There can be other parameters for this detectMultiscale function which can be used anytime based on the need of expected result.

So, using this detectMultiScale function, WE can find faces in the image and this function also represents the positions of the face in the image. After finding the positions of the face using the openCV built in rectangle function, a rectangle has been drawn on to the original coloured image. Finally, the algorithm returns the image to verify if the detection is correct and find the real face in the image. Here, WE use matplotlib to convert the gray image into its original BGR colour.

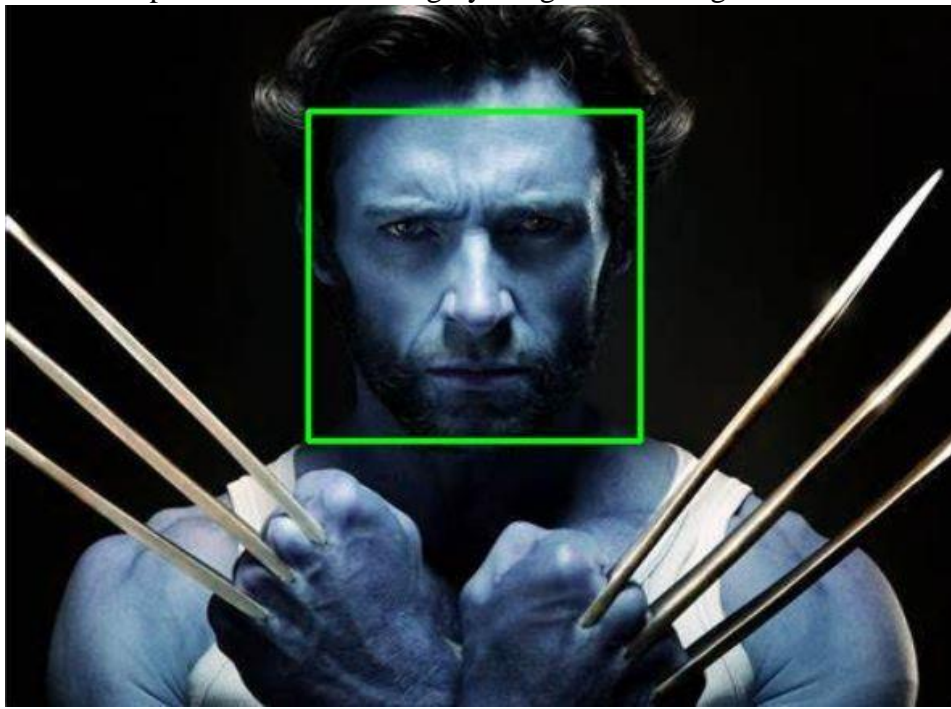

*Figure 19: Detected Face*

Though this detectMultiScale function is very much helpful to detect a face, to make this reusable to all the faces in an image WE use one more function which is called "detect_faces". It also takes 3 input parameters- loaded CascadeClassifier, the test image and the scale factor. Inside the function, WE made a copy of the original image and converted it to the grayscale, so that all the operations of the face detectors can be done on that copy grayscale image. Then, WE called the detectMultiScale built in function from the Cascade Classifier to provide the list of detected faces as previously discussed. Actually, WE repeat the full process as described earlier, just WE did it inside this detect_faces function, so the function can be reusable. WE detect all the faces from a group image using this function with the help of HAAR cascade classifier.

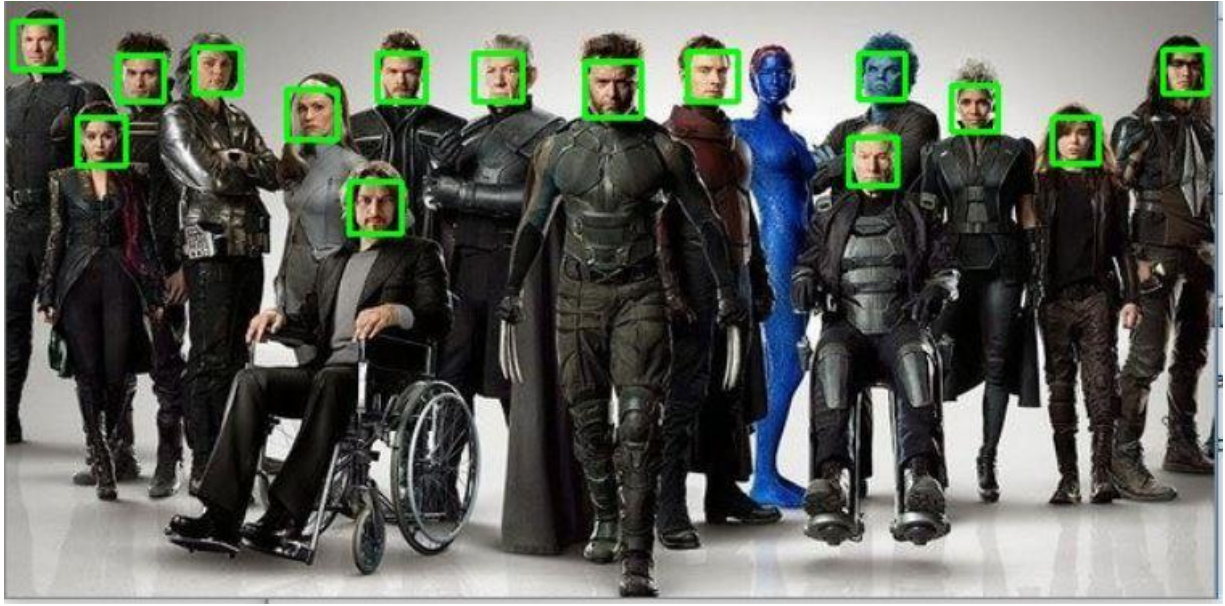
*Figure 20: Detected Face from a group image using HAAR cascade*

## LBP Cascade Classifier Using

As we are using openCV the coding is much easier than ever. To run the algorithm for LBP cascade classifier simply WE just replace the HAAR cascade classifier xml file with the LBP cascade classifier file and loading this classifier detects the faces as well. For implementation, WE did not do any change in coding.

As the discussion about face detection implantation from image is complete, now WE move on to the next part of face detection which includes detection from live video streaming.

# Face Detection from Live Video Streaming

Previously, we did face detection from image where we do not need any hardware, but when we are talking about live video detection it goes without saying that we need a camera for this. This is the only part of this project where we need a hardware device. It can be an external webcam for desktop device or the built-in webcam of a laptop.

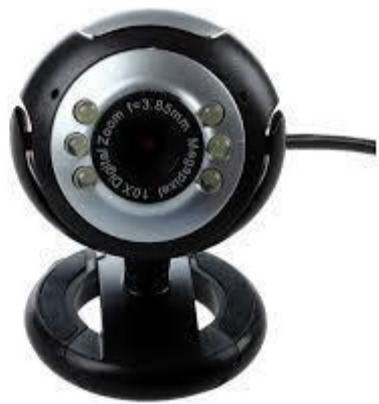

*Figure 21: Webcam (Hardware device) for face detection from live video streaming.*

To implement the algorithm, here, WE used HAAR cascade frontalface_detect classifier and HAAR cascade eye_detect classifier. Then WE import cv2 and numpy in the algorithm like previously WE used for face detection from image. WE captured the video using the webcam and then the classifier read that image. After that, using the same detectMultiscale function, the algorithm found faces in the live video streaming. The code was very simple, which is shown in below image.

```python
import numpy as np
import cv2

face_cascade = cv2.CascadeClassifier('open cv file/haarcascade_frontalface_default.xml')
eye_cascade = cv2.CascadeClassifier('open cv file/haarcascade_eye.xml')

cap = cv2.VideoCapture(0)

while 1:
    ret, img = cap.read()
    gray = cv2.cvtColor(img, cv2.COLOR_BGR2GRAY)
    faces = face_cascade.detectMultiScale(gray, 1.3, 5)

    for (x,y,w,h) in faces:
        cv2.rectangle(img,(x,y),(x+w,y+h),(255,0,0),2)
        roi_gray = gray[y:y+h, x:x+w]
        roi_color = img[y:y+h, x:x+w]

        eyes = eye_cascade.detectMultiScale(roi_gray)
        for (ex,ey,ew,eh) in eyes:
            cv2.rectangle(roi_color,(ex,ey),(ex+ew,ey+eh),(0,255,0),2)

    cv2.imshow('img',img)
    k = cv2.waitKey(30) & 0xff
    if k == 27:
        break

cap.release()
cv2.destroyAllWindows()
```

*Figure 22: Code for face detection from live video streaming*

This simple code helps to detect the face from the live video streaming. This can also find the face in a moving position. For example, If WE move our heads constantly this can detect the face without any discontinuation. Moreover, If WE have some more images in our hand, it can detect the face of that image too, though sometimes it cannot detect all the faces from the image which WE have in our hand because of the lighting coming from the laptop, which makes the image in our hand blur. The face detection live result is shown in below image.

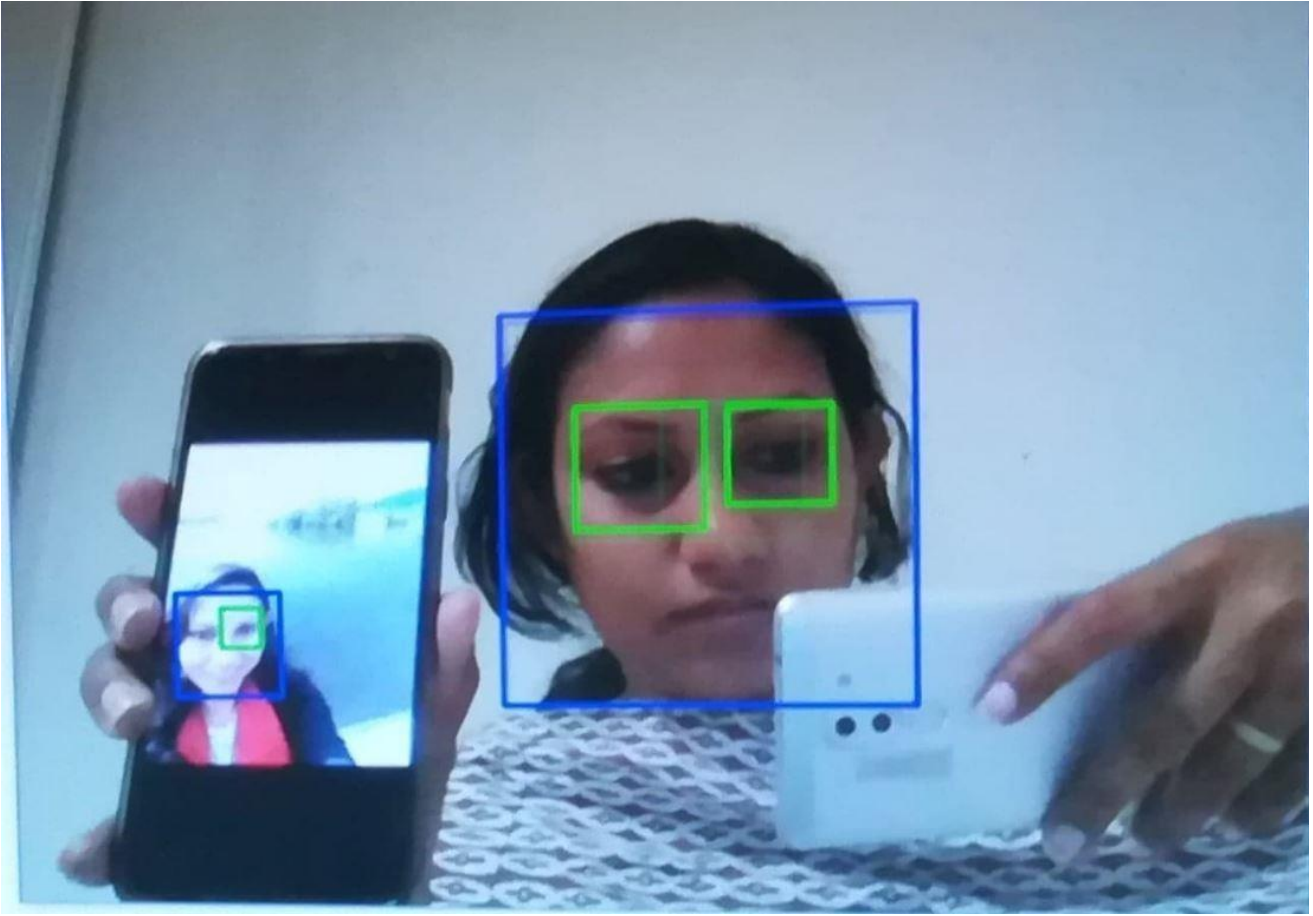
*Figure 23: Face detection result from live video streaming*

From the above picture, we can see that this algorithm can detect all the two faces from the live video streaming. Though, WE have another phone in our other hand which covers a small part of our face, still this algorithm can detect our face, as it detected our eye areas. For this reason, beside using the HAARcascade face detection classifier, WE also use the eye cascade classifier to get the more accurate result as this detection is happening on to a moving object.

# Face Recognition Implementation

## LBP Cascade Classifier Using

Before WE discussed, the face detection implementation process describing HAAR cascade classifier, now for face recognition WE will explain the process for LBP histogram cascade classifier.

This face recognition process is also linked with the face detection process, WE just discussed earlier. So, after detecting a face when it comes to recognizing the face, then the first things come to our brain is to know about that face; and to know about that face we need to train our brain about that face. Similarly, for face recognition, at first, we need to train our system about the face so that it can recognize the face in future. For implementing our system, we need to prepare training data for the face recognition. As we are using openCV so, by this time we already know that openCV face recognizer accepts data in a specific format which consists with two vectors. One of them is used for faces of all the persons. Another one is used as a specific label for each of the face, which is used by the recognizer, to memorize the face is belong to which specific person. Now WE move on to train the recognizer. For this, WE used a satisfactory number of images of two different actor. Training of the recognizer would be better depending on the number of the image. WE took a lot of different image of same person in our implementation, so that, the recognizer can learn a lot about the different facial expression of that person. For example- smiling, sad, happy, with beard, without beard, with hair, without hair etc. WE took 10 different images of each of the actors "Hugh Jackman" and "James McAvoy" as training data. WE create a folder under our face recognition project folder and named it "training_data". Here WE create 2 more folder named 's1' and 's2' where 's1' and s2 is the integer label assigned to the actors. Folder s1 contains training image of 'Hugh Jackman' and folder s2 contains all the training image for 'James McAvoy'. The directory structure shown in below image.

```
training-data
|-------------- s1
|               |-- 1.jpg
|               |-- ...
|               |-- 12.jpg
|-------------- s2
|               |-- 1.jpg
|               |-- ...
|               |-- 12.jpg
```

Figure 24: Training data directory

**Training Data Preparation**

The training data preparation exactly follows four steps. They are discussed below in points.

- Inside the training_data folder we create sub folder s1 and s2 for each of the actors where we put their images to their dedicated folder. While training the data one step is to read that folder names.
- The sub folder names follow the format s1 and s2, where s is the subject and ½ is an integer which represents the label. This label number is clarifying which label is dedicated to which subject. For example, s1 means the subject in this folder has label 1 and the s2 means the subject in this folder has label 2. In this step, we extract the integers which we will use in next step.
- From the subject folder, read all the images and detect face from each of the image
- Add each face, to faces vector with corresponding person label which was extracted in previous step.

After following these steps produced face and label vectors are shown below:

| FACES | LABELS |
|---|---|
| person1_img1_face | 1 |
| person1_img2_face | 1 |
| person2_img1_face | 2 |
| person2_img2_face | 2 |

*Figure 25: Face and label vectors*

In the coding part WE used the LBP cascade classifier for detecting the face. After that, for preparing the training data WE defined a function named "prepare_training_data". This function takes the path as a parameter where WE stored the training_data subjects. This function follow that 4 steps of data preparation stated blow.

In step 1, to read the names of all folders which are stored with training data, WE defined a method named "os.listdr" which takes the path as a parameter. Then WE defined the labels and face vectors.

In step 2, the algorithm extracts the label information from each subject folder name, because the folder name provides the label information of each subject.

In step 3, the algorithm read the name of all images of the current subject, and move that images one by one. Then, using openCV's built in "imshow" and "waitkey" method, the algorithm showed the current image which is being moved and temporary halt the code flow for 100ms interval. Then, the algorithm implements the face detection code on the current picture to detect the face from the image.

In step 4, the algorithm adds the detected face and label with its respective vectors. **Train Face Recognizer**

After the training data preparation, the next step is to train the face recognizer with that data. In the research methodology section, we already discussed about the methods of face recognition provided by openCV. Here, WE are going to use LBPHFaceRecognizer. The most interesting part is that the coding will remain the same, WE will just change the recognizer and will see the results.

The algorithm already prepares the train data and now initialize the face recognizer with "face_recognizer = cv2.face.createLBPHFaceRecognizer()" this code. Now, the next part is to train the recognizer. For this, WE use a method called "train" where WE pass the faces-vector and labels-vector as parameter. Here, before passing the label vector WE convert the value into NumPy array as OpenCV expects the value into a shape of an array.

**Predict Face**

Now, the final part is to predict the faces as the training data is already prepared, and the recognizer is ready and trained. This is the crucial part where we will be able to find out if the algorithm used is actually recognizing the trained faces or not. For this, two test images is selected from the movie actor faces which WE already give training to the recognizer. After detecting the face from the test image and then using our train recognizer method, WE pass those faces to see if it can predict the face. For this, WE create one function which WE named as "predict", where WE called the method for face recognizer to test the recognizer on the given test images. Then, WE called two utility OpenCV built in function inside the predict funtion, one is "draw_rectangle" to draw rectangle box around the face and another is "draw_text" to write the name of the actor near the rectangle box. Finally, after defining the "predict" function successfully WE called the function on the test images and showed the result that if the face recognizer accurately can predict the face of the test images or not. The result WE found from LBPHFace recognizer is shown in below picture.

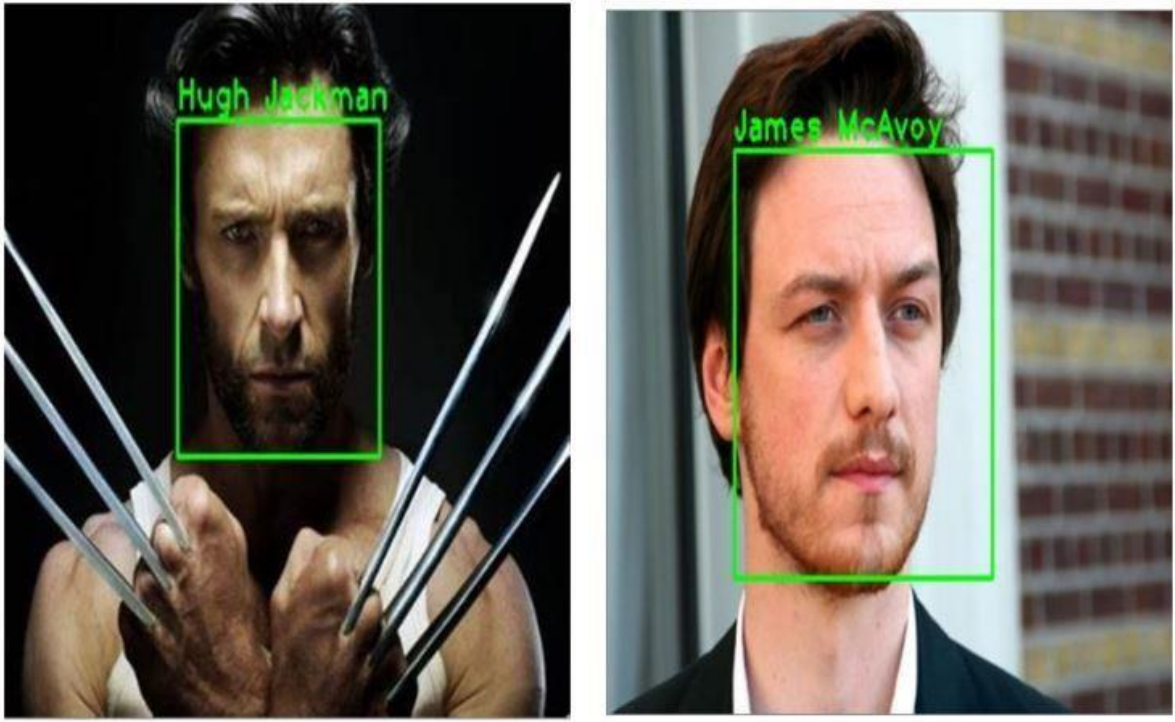

*Figure 22: Face recognition using LBPH face recognizer*

So, the above picture is showing that the algorithm successfully recognizes the faces using the LBPH face recognizer algorithm from OpenCV. Though, this result WE have achieved using LBPH face recognizer but as discussed earlier just changing one line of code, we can use any of the face recognizer with the same algorithm. In next section WE will analyse the test results which WE have achieved using different recognizer.

# Test result Analysis

After completing the system implementation, the next necessary part is to evaluate all the test results which we have achieved using all of the techniques for face detection and recognition. From the evaluation of the test results, we will be able to find out which technique is faster, which technique provides more accuracy rate and depending on that, we can select the best technique while considering for which situation we are using the algorithm.

# Face Detection Speed and Accuracy Test

As we know OpenCV provides two classifiers for face detection, WE use both of them in the same algorithm to see which one is faster. To find out the calculating speed, WE use a group image which is shown below:

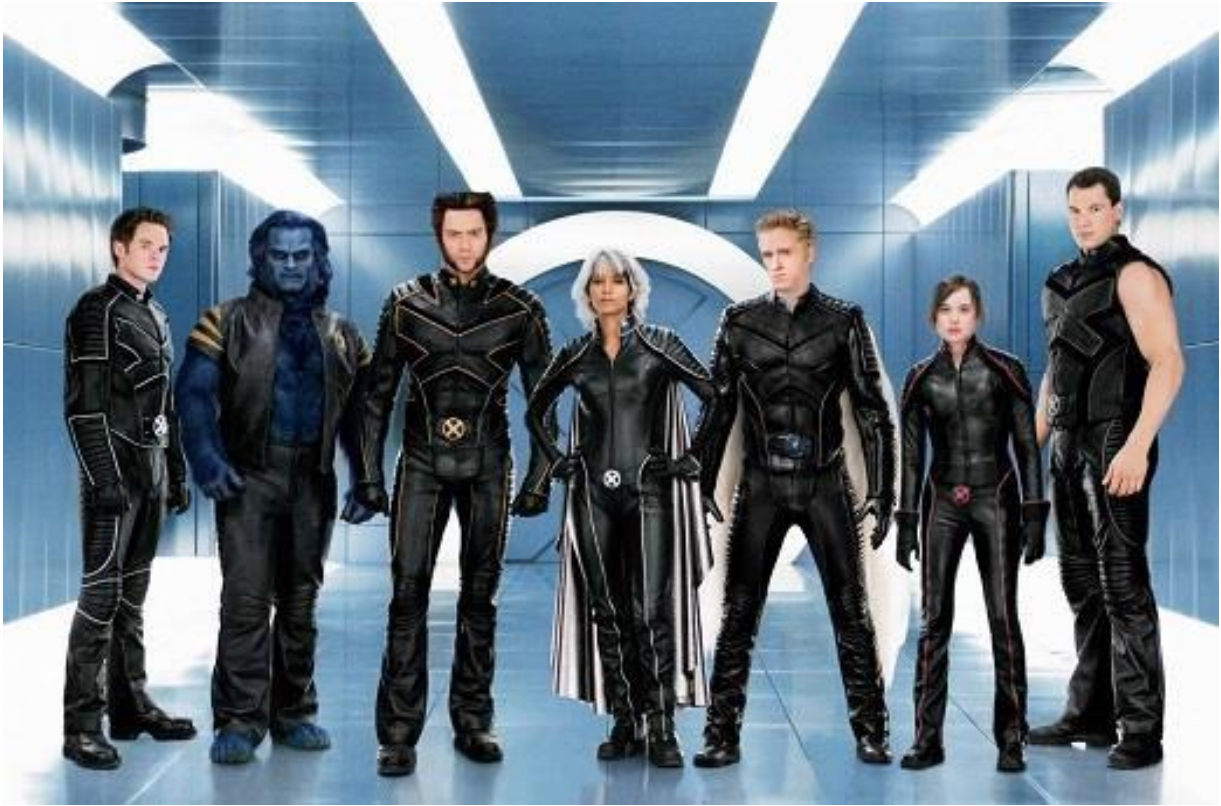

Figure 27: Test image for calculating speed

After that, WE write a code for detecting the time delay. First, WE load both the classifier and the test image. Then, WE used one of the library function provided by python which is time.time() to keep track of time. Before, WE start detecting face on that test image, WE take the input for start time t1 and then WE called the function for face detection and finally WE defined the end time t2.

Then WE calculate the difference between start time and end time, which WE initialize as dt1. The result of dt1 is the time the classifier takes to complete the detection.

The same image and same code WE used to test both classifiers. The result with HAAR cascade classifier is shown in below image.

```
Python 3.5.0 (v3.5.0:374f501f4567, Sep 13 2015, 02:16:59) [MSC v.1900 32 bit (In
tel)] on win32
Type "copyright", "credits" or "license()" for more information.
>>>
 RESTART: D:\Software\face recognition software\face detection and recognition\T
est image face detection of group pic-haarcascade.py
Faces found:  7
HAAR cascade detection time 0.06791877746582031
>>> 
```

Figure 28: HAAR cascade detection time

From the above picture we can see that HAAR cascade classifier finds all the faces within 0.07 sec. and the accuracy rate is also 100%. The detected faces are shown in below image.

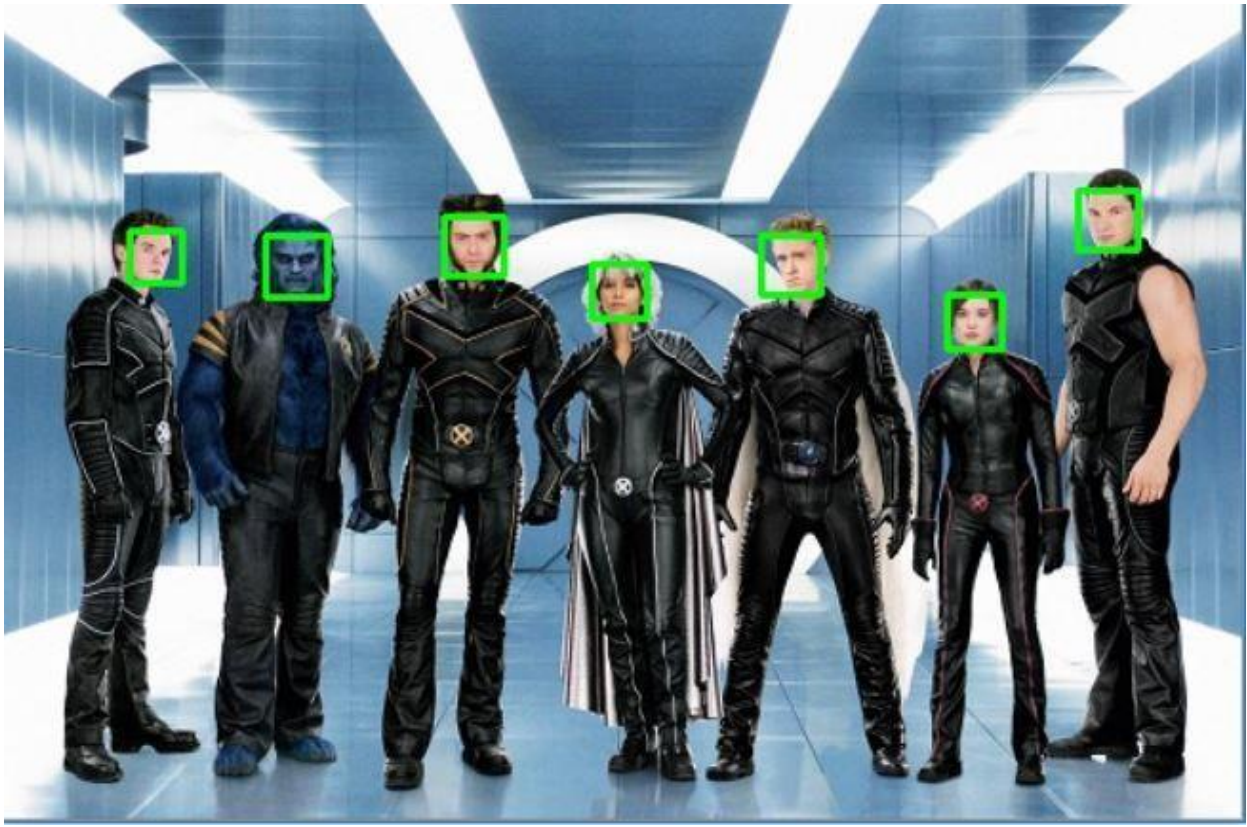

*Figure 29: All the faces detected by HAAR cascade classifier*

From the above image we can see, though one of the faces is wearing a mask, still it is being detected by the algorithm. That is because all of the faces here are frontal faces and close to the camera. If a face is far from the camera or wearing mask sometimes it is difficult to find that face, which is discussed later on in this report. Now, WE use the same code for LBPH cascade classifier and see the result.

```
Python 3.5.0 (v3.5.0:374f501f4567, Sep 13 2015, 02:16:59) [MSC v.1900 32 bit (In
tel)] on win32
Type "copyright", "credits" or "license()" for more information.
>>>
 RESTART: D:\Software\face recognition software\face detection and recognition\T
est image face detection of group pic-lbp.py
Faces found:  5
LBP cascade classifier detection time 0.02931070327758789
>>> 
```

*Figure 30: LBP cascade classifier detection time*

The above picture shown that the face detection time for LBP is lower than Haar cascade classifier which is only 0.03 second but it detects 5 faces. So, the accuracy is not as higher as HAAR cascade classifier. The accuracy result for detecting number of positive faces is shown in below images.

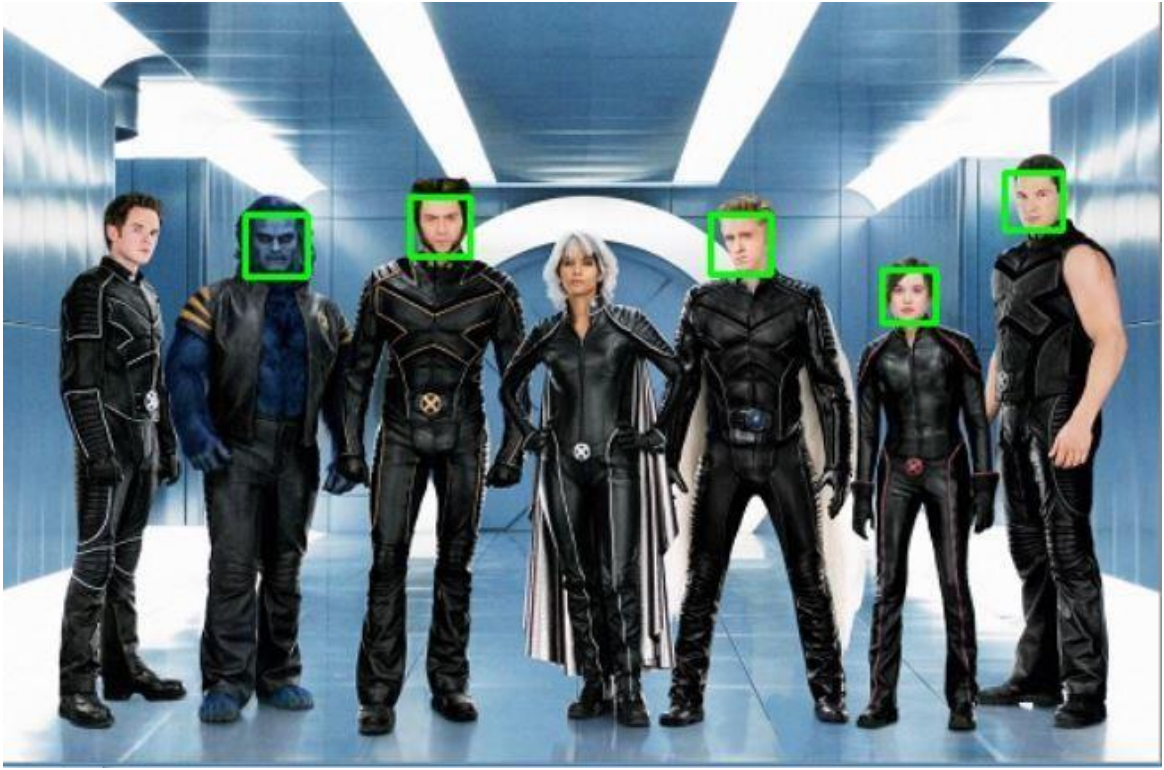

*Figure 31: Five faces detected by HAAR cascade classifier*

From the above image, we can see that 5 faces out of 7 faces has been detected, which proves the accuracy rate almost 72% but it is 0.04 sec faster than Haar cascade classifier.

## Challenges Involved with Accuracy Rate

Above WE showed that, in case of accuracy HAAR cascade classifier is more reliable than LBP classifier, which is actually true, but in some cases it is hard to find out all the actual faces from the test image which is because of the slight changes in images. It can happen if one of the faces is far from the camera or issue with lighting. In this project, WE also experiment this with another test image which is shown below.

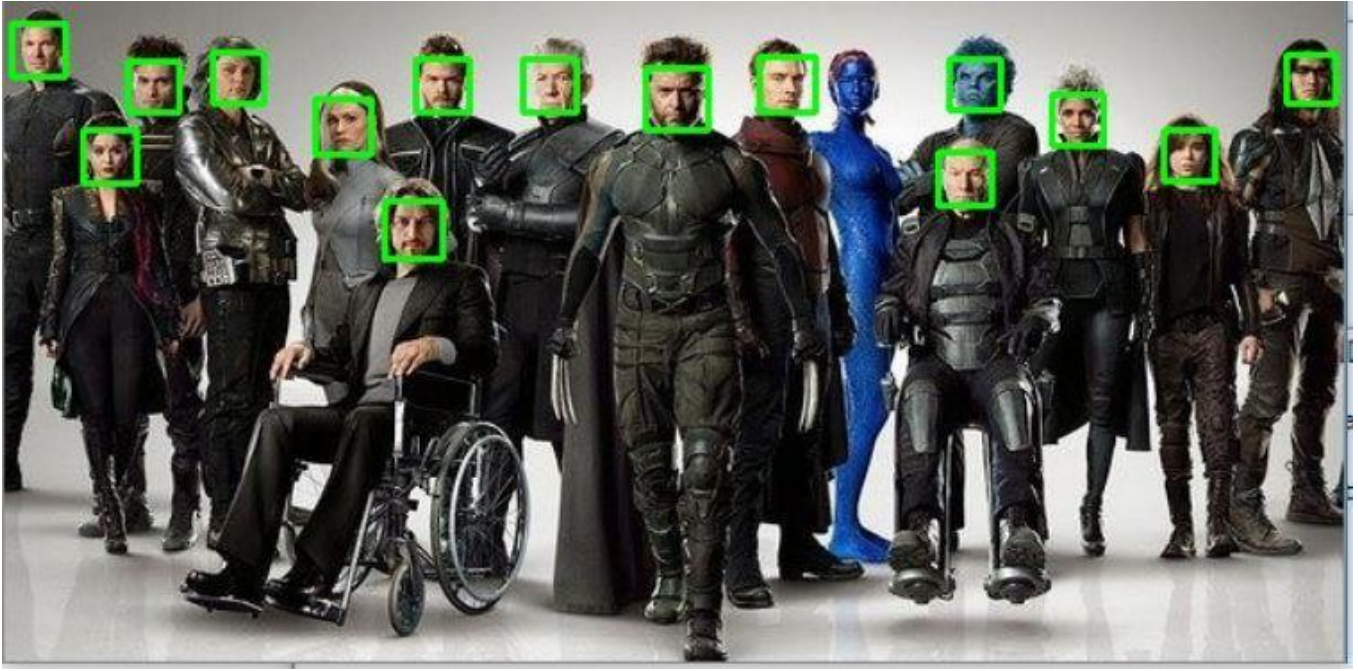
*Figure 32: Fifteen faces detected by HAAR cascade classifier*

From the above image we can see that, within 16 faces 15 faces has successfully detected by the HAAR cascade classifier. The face which is wearing a musk cannot be detected by the classifier. This is because the pattern is different from other faces as it is wearing a mask, and the face is far from the camera as well. Moreover, if you look at the face with deep concentration, you can be able to see that, the left side of the face is darker from the right side of the face, which makes it difficult for the HAAR cascade classifier to detect the face as it cannot be able to compare the distance between the eyes, nose etc. and also cannot be able to determine the principal component of that face.

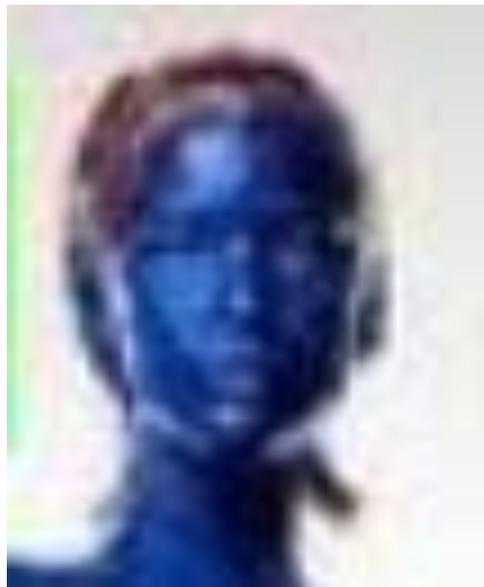
*Figure 33: Face detection failed for the blur image*

The above picture WE cropped from the group picture, to show that the facial features of this image is not clear enough to detect by the recognizer as it is not enough trained for this type of face.

Though the algorithm could not find all the faces but still its accuracy rate for faces with principal component is admirable and the accuracy rate is almost 94% using the HAAR cascade classifier.

The same image we used for detection using the LBPH cascade classifier using the same algorithm. The result shown in below picture.

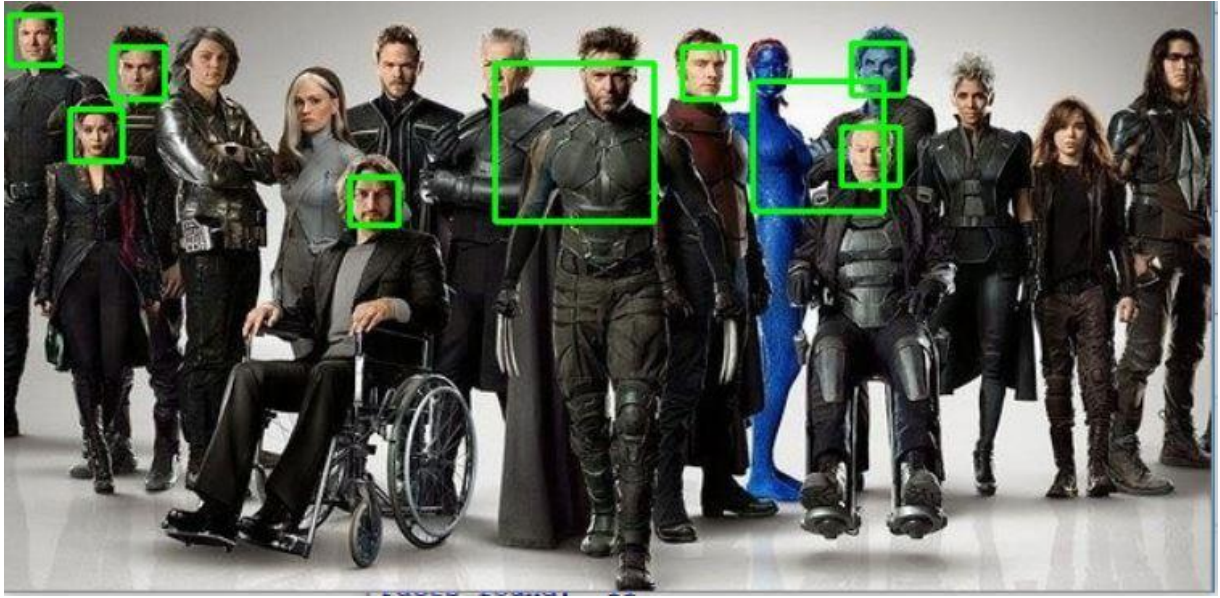
*Figure 34: Seven faces detected by LBPH cascade classifier*

The above image shown represents, seven detected faces and two false positive faces which represents that using LBPH cascade classifier, here the accuracy rate is almost 44%, that means less than 50%. The false positive result may be happened because of change of the lights and shadow in the image.

So, it can be said that to get 100% accurate result is not always possible but to get the most possible accurate result HAAR cascade classifier can be used, though it takes a bit more time than LBPH cascade classifier.

## Face recognition testing

Now, WE move on to testing our face recognizers. As we see in previous section, we use LBPH face recognizer to recognize the faces. Below the picture with the results achieved using LBPH recognizer.

```
Python 3.5.0 (v3.5.0:374f501f4567, Sep 13 2015, 02:16:59) [MSC v.1900 32 bit (In
tel)] on win32
Type "copyright", "credits" or "license()" for more information.
>>>
 RESTART: D:\Software\face recognition software\face detection and recognition\F
ace recognition with LBPH.py
Preparing data...
Data prepared
Total faces:  6
Total labels:  6
Predicting images...
Prediction complete
>>> 
```

*Figure 35: Face recognition result using LBPH face recognizer*

From the above picture we can see that our used algorithm, prepare the training data successfully. Then it counts the number of trained face and trained labels for those faces, which is 6. Then it starts predicting the image and when the algorithm successfully recognizes the faces it shows "prediction complete" and return the images as well, which is shown in below picture.

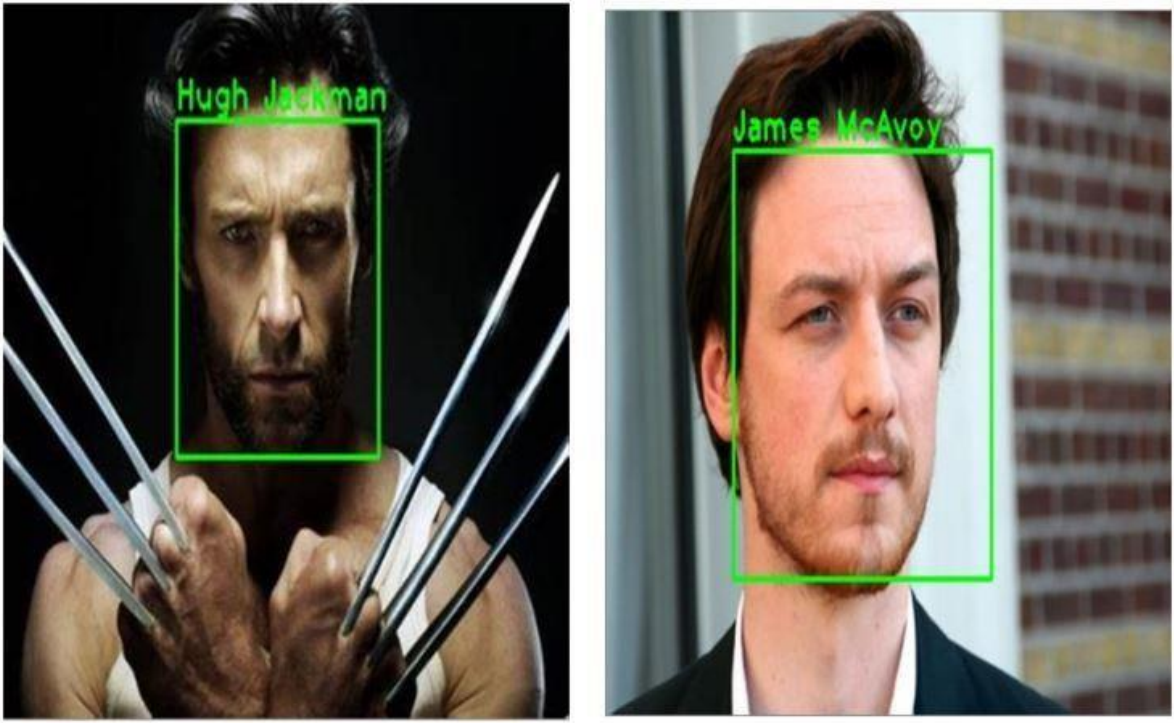

*Figure 33: Face recognized successfully LBPH face recognizer*

The above image shown that the faces are successfully recognized, and their names are also written on top of the face box. So, it is easy to recognize the faces using the LBPH recognizer.

Then, WE use the same algorithm to test the other two, eigenfaces recognizer and fisherfaces recognizer by changing that one-line recognizer initialization code in our algorithm and the result WE have got is shown in below image.

```
RESTART: D:\Software\face recognition software\face detection and recognition\F
ace recognition with LBPH.py
Preparing data...
Data prepared
Total faces:  6
Total labels:  6
Traceback (most recent call last):
  File "D:\Software\face recognition software\face detection and recognition\Fac
e recognition with LBPH.py", line 262, in <module>
    face_recognizer.train(faces, np.array(labels))
cv2.error: OpenCV(4.1.0) C:\projects\opencv-python\opencv_contrib\modules\face\s
rc\eigen_faces.cpp:72: error: (-210:Unsupported format or combination of formats
) In the Eigenfaces method all input samples (training images) must be of equal
size! Expected 27889 pixels, but was 6724 pixels. in function 'cv::face::Eigenfa
ces::train'
```

*Figure 37: Face recognition result using eigenface face recognizer*

The above image shown that, the algorithm works successfully up to detecting the training face and their labels, but the recognizer failed to train itself. The error is showing that to use the eigenface method, all the train images must have to be equal size. Here, WE used images which are different in size from one another. Also, it defines the expected size is within 27889 pixels whereas our training images pixel size was 6724 pixels. So, what we can see from that test result is, in Eigenfaces method it is important to maintain the image size and pixels to conquer the accurate recognition result. WE tried the same with the Fisherface recognizer and it also shows the same test result as eigenfaces. So, the most convenient way to recognize a face is to LBPH face recognizer as this recognizer did not give us any size or pixel limit for the images and easily trained the recognizer with the given image, also successfully recognize the faces. However, converting each training and test image into same size and within 27889 pixels, the accurate result is achieved like LBPH face recognizer. OpenCv also has an built in function which is cv.resize() to resize any image to expected size. WE used this function to achieve the correct result.

After finding and analysing the test results, next section WE will show the comparison within the face detection and recognition techniques and will provide a decision about which technique is more efficient.

# Efficiency Comparison

## Comparison within the face detection classifier

In this paper, WE showed two classifiers for face detection. WE used both of them for face detection and showed the test result in previous section. From the literature review, system implementation and test result analysis WE found some advantages and disadvantages of both classifiers which is shown in below table:

| Classifier Algorithm | Advantage | Disadvantage |
|---|---|---|
| HAAR Cascade Classifier | 1. High face detection accuracy<br>2. Amount of false positive is very less | 1. Comparably complex computation<br>2. Slow performance<br>3. Less accuracy rate if the faces are black<br>4. Hard to find faces in different lighting conditions<br>5. Not very much robust to occlusion<br>6. HAAR does all the computation in floats which is very difficult for an embedded system. |
| Local Binary Pattern Cascade Classifier | 1. Simple computation system<br>2. Fast performance<br>3. Take less time while training the faces<br>4. Can detect local illumination changes<br>5. Robust to occlusion than HAAR.<br>6. LBP does all the calculations in integer which is useful for an embedded system. | 1. Less face detection accuracy rate<br>2. Amount of false negative is high |

*Table 1: Comparison between HAAR cascade classifier and LBP classifier*

## Comparison Within the Face Recognition Algorithm

We discussed three of the face recognizer algorithms in this paper. WE found out the efficiency of the algorithm while implementing them, analysing their test results and also from the literature review which WE have researched before starting our implementation. Below is some difference between principal component analysis which is used by eigenfaces and Linear Discriminate analysis which is used by fisherfaces.

| Principal Component Analysis-Eigenface | Linear Discriminate Analysis-Fisherface |
|---|---|
| 1. This analysis observes the greatest variance.<br>2. It maximizes the variation<br>3. Differentiation between feature classes are not accurate<br>4. Take more space and time<br>5. Suitable for representing set of data<br>6. The first component of PCA is shown in image (Dinalankara, Lahiru. 2017)<br>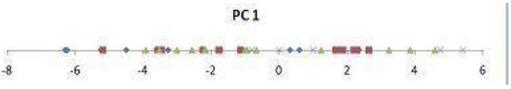 | 1. It observes an interesting dimension so that it can maximizes the differences between the mean of the normalised classes. [21]<br>2. It minimises the variation and maximizes the mean distance between different classes.<br>3. Differentiation between features are better than PCA<br>4. Take less space and fast<br>5. Suitable for classification<br>6. The first component of LDA is shown in image (Dinalankara, Lahiru. 2017)<br>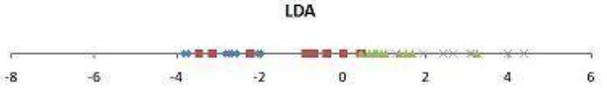 |

*Table 2: Comparison between PCA and LDA*

These are some important difference shown in above table but the analysis we found while implementing is described in detail below.

From this implementation, it has been found, to get the accurate face recognition in case of eigenface recognition algorithm, the training data set should be bigger. It cannot predict face when there was less training data, increasing the number of training data, made the wrong recognition accurate, though it did not help to recognize more subject. Not only this, but also this algorithm has some limitations that we already discussed in our result analysis section, that including test images, all the training images should be in same size and the pixel value is also defined by this algorithm.

In fisherface algorithm, we found the same result as eigenface algorithm in our implementation and also the limitations with image size and pixel was the same as eigenfaces algorithm. However, from some literature review we found out that, fisherface algorithm behaves quite differently with the change of number of training images. Sometimes, it recognizes the correct picture with 20 training image and sometimes increasing the number of training image the result is same or worse.

The best algorithm for recognition was LBP from our implementation, as it recognized the trained faces from the test image within the shortest time and also it was accurate. Moreover, there were no dependencies for size limit, like eigenface and fisherface algorithm. WE used any size of picture and though each of them was not in same size, but still it works fine. From the literature review, it is also found that this algorithm is broadly used to recognize the face because of the accuracy rate.

# Error Handling

Before resizing the picture, the results with eigenface and fisherface was unsuccessful as only LBPH can recognize the faces. However, we design another algorithm which is described in researched methodology section followed by a flow chart which helps to avoid this problem and result was successful after using that. Another algorithm which we used to define the threshold also helps to determine the faces with distances which are not been recognized and also helped to pass the face recognition algorithm by returning -1 value. Algorithm discussed in research methodology section helps to solve the problem where people tilting their head. This was fixed by the algorithm by identifying the locations of the eyes and rotating the image. Another problem was that the brightness of the images were different somewhere. It resolves by using the same algorithm and make the brightness average which improves the performance a lot.

# Conclusion

This article discusses a fundamental project for face identification and recognition utilizing photos of movie actors, which is essentially a research-based endeavor. This article introduces the basic concept of face detection and identification and then implements it using only the classifiers provided by OpenCV. It analysed the findings of similar initiatives and then demonstrated the fundamental concepts of face detection and recognition in the part on the research technique. Through extensive research and execution, it has been shown that the HAAR cascade classifier is the most effective for face detection. Face detection in live video streaming was likewise satisfactory with this classifier, even when the subject's head moved concurrently. Though we used three recognizers for face recognition and obtained successful results, when all variables are considered, the Local Binary Pattern paired with HAAR cascades is the best solution for face recognition discovered in this experimental project that can be implemented as a complete system. Additionally, it will be a cost-effective system that may be used for automatic attendance tracking, security systems in autos, and ATMs, among other applications.

This article provides an overview of the whole face detection and identification system for a beginner, but advanced implementation can be accomplished by combining a few other approaches with this information. In the future, there is a scope to use this face detection and identification approach to do research on wound detection and recognition, finalizing the necessary therapy for that wound, which will have a significant impact on the health care system.